\documentclass[10pt,twocolumn,letterpaper]{article}

\usepackage{cvpr}              % To produce the CAMERA-READY version
\renewcommand{\paragraph}[1]{\vspace{0.3em}\noindent\textbf{#1}}

\definecolor{cvprblue}{rgb}{0.21,0.49,0.74}
\usepackage[pagebackref,breaklinks,colorlinks,allcolors=cvprblue]{hyperref}

\def\paperID{14520} % *** Enter the Paper ID here
\def\confName{CVPR}
\def\confYear{2026}

\usepackage{xspace}
\def\method{OCH3R\xspace}

\title{OCH3R: Object-Centric Holistic 3D Reconstruction}

\author{Yi Du \quad Yang You\quad Xiang Wan \quad Leonidas Guibas\\
Stanford University\\
{\tt\small \{duyi, yangyou, oscarwan, guibas\}@stanford.edu}
}

\begin{document}
% \maketitle

\twocolumn[{%
\renewcommand\twocolumn[1][]{#1}%
\maketitle
\vspace{-1em}
\centering
\includegraphics[width=.85\linewidth]{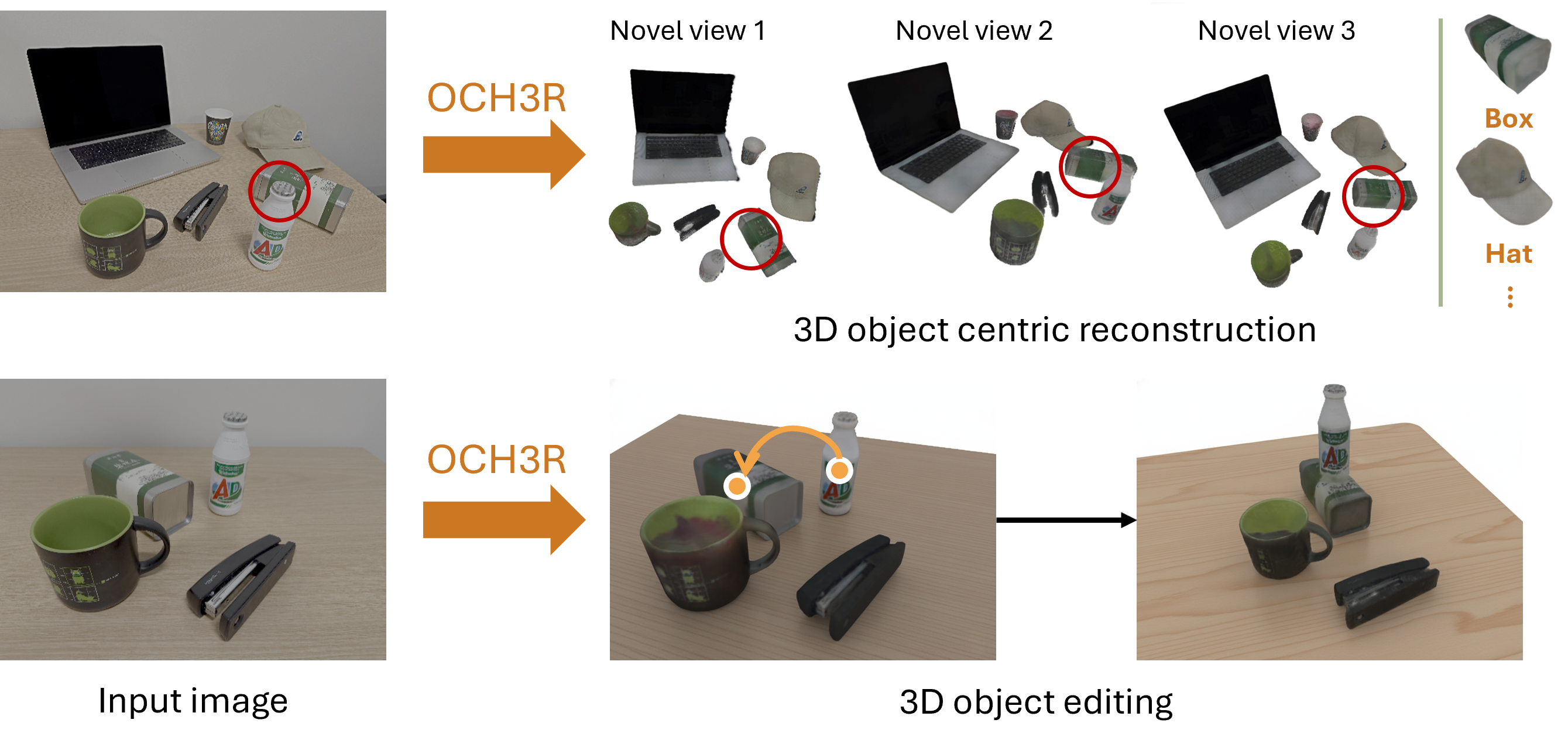}
\captionof{figure}{\textbf{\method enables fully object-centric 3D scene reconstruction from a single RGB image.}
Given one input view, \method discovers all object instances, predicts their 6D poses, 
and reconstructs each object as a manipulable 3D Gaussian model in a single forward pass. 
Our feed-forward, per-pixel prediction framework supports selecting, moving, and rendering objects from arbitrary novel views without external segmentors or multi-stage pipelines. 
\method produces amodally complete, editable 3D objects and generalizes to cluttered real scenes, enabling downstream tasks such as rearrangement and AR editing. The red circles highlight occluded regions that \method successfully completes in 3D.\vspace{1em}}
\label{fig:teaser}
}]

\begin{abstract}
Object-centric scene understanding is a fundamental challenge in computer vision. Existing approaches often rely on multi-stage pipelines that first apply pre-trained segmentors to extract individual objects, followed by per-object 3D reconstruction. Such methods are computationally expensive, fragile to segmentation errors, and scale poorly with scene complexity. We introduce \textbf{OCH3R}, a unified framework for  \underline{\textbf{O}}bject-\underline{\textbf{C}}entric 
\underline{\textbf{H}}olistic
\underline{\textbf{3}}D \underline{\textbf{R}}econstruction from a single RGB image. \method performs one forward pass to simultaneously predict all object instances with their 6D poses and detailed 3D reconstructions. The key idea is a transformer architecture that predicts per-pixel attributes, including CLIP-based category embeddings, metric depth, normalized object coordinates (NOCS), and a fixed number of 3D Gaussians representing each object. To supervise these Gaussian reconstructions, we transform them into canonical space using the predicted 6D poses and align them with pre-rendered canonical ground truth, avoiding costly per-image Gaussian label generation.
On standard indoor benchmarks, \method achieves state-of-the-art performance across monocular depth estimation, open-vocabulary semantic segmentation, and RGB-only category-level 6D pose estimation, while producing high-fidelity, editable per-object reconstructions. Crucially, inference is fully feed-forward and scales independently of the number of objects, offering orders-of-magnitude speedups over conventional multi-stage pipelines in cluttered scenes.
\end{abstract}   
\vspace{-1em}
\section{Introduction}
\label{sec:intro}

Understanding a scene as a composition of discrete, posed objects from a single image is a long-standing goal in computer vision. Many downstream applications including robotic manipulation, AR editing, and simulation rely on object-centric outputs~\cite{yao2025castcomponentaligned3dscene}, where each object is represented with geometry, pose, and semantics that can be selected or manipulated. We study the following problem: given a single RGB image of an indoor tabletop scene, recover all objects together with their 6D poses and corresponding 3D Gaussians in one forward pass.

Prior work largely falls into two groups. Scene-level, feed-forward methods (\eg, one-pass Gaussian predictors)~\cite{charatan2024pixelsplat3dgaussiansplats, zhang2024gslrmlargereconstructionmodel, szymanowicz2025flash3dfeedforwardgeneralisable3d, xu2025depthsplatconnectinggaussiansplatting, ye2024poseproblemsurprisinglysimple} are fast and photorealistic, but typically produce an undifferentiated ``soup'' of geometry without instance-level structure, canonical frames, or object poses. Object-level approaches~\cite{yao2025castcomponentaligned3dscene, irshad2022shapoimplicitrepresentationsmultiobject, irshad2022centersnapsingleshotmultiobject3d} instead rely on multi-stage pipelines that begin with external open-vocabulary segmentors and then perform per-object reconstruction, alignment, and correction. These systems often depend on RGB-D inputs or category-specific priors, and they are fragile to upstream errors, difficult to scale with the number of objects, and not trained end-to-end. As a result, their robustness and accuracy degrade in cluttered tabletop scenes.

To address these limitations, we introduce \textbf{OCH3R}, a unified, object-centric, holistic 3D reconstructor that converts a single RGB image into a set of posed 3D objects in one pass. The key in our design is a 48-layer transformer that predicts dense, pixel-aligned attributes: CLIP \cite{radford2021learningtransferablevisualmodels}-based category embeddings, metric depth, normalized object coordinates (NOCS)~\cite{wang2019normalized}, and a small set of 3D Gaussians~\cite{kerbl20233d} per pixel. During inference, object instances and their poses are recovered by clustering the semantic embeddings and estimating each object's $\mathrm{SIM}(3)$ pose using the predicted NOCS field.

To train the Gaussian representation, we allow Gaussians at each pixel to move freely off the pixel rays to compensate for (self-) occlusion. Rather than supervising Gaussians per training image~\cite{szymanowicz2024splatter,szymanowicz2025flash3d}, we adopt Canonical-Space Supervision: per-object Gaussians are transformed into canonical space using its $\mathrm{SIM}(3)$ pose, and their renderings are optimized against pre-rendered ground truth in the canonical frame. This eliminates the need for costly per-image Gaussian labels and promotes amodal shape completion.

We train on a curated, large-scale dataset that integrates PACE \cite{you2024pacelargescaledatasetpose}, Omni6DPose \cite{zhang2024omni6dposebenchmarkmodeluniversal}, GSO \cite{downs2022googlescannedobjectshighquality}, and Hypersim \cite{roberts2021hypersimphotorealisticsyntheticdataset}, offering broad coverage across object categories, poses, and occlusion patterns.

Our experiments show that \method substantially outperforms previous baselines across all evaluated tabletop object-centric benchmarks.
\method delivers consistently higher geometric accuracy, better semantic alignment, and significantly more complete amodal reconstructions. Importantly, because \method reconstructs all objects in a single forward pass, it achieves orders-of-magnitude faster inference than multi-stage pipelines while avoiding their brittleness to segmentation or pose-estimation errors. Together, these results highlight the effectiveness of our unified formulation and its practical advantages for real-world object-centric applications.

To summarize, our contributions are as follows:
\begin{enumerate}
\item We construct a large scale dataset for holistic object centric 3D scene representation. We assemble, relabel, and align PACE \cite{you2024pacelargescaledatasetpose}, Omni6DPose \cite{zhang2024omni6dposebenchmarkmodeluniversal}, GSO \cite{downs2022googlescannedobjectshighquality}, and Hypersim \cite{roberts2021hypersimphotorealisticsyntheticdataset} into a unified dataset designed for object centric 3D tasks, providing per instance masks, segmentation labels, $\mathrm{SIM}(3)$ poses, and 3D models.

\item We propose a model that yields high-fidelity 3D reconstructions, recovering fine-grained geometry and amodal structure, while jointly predicting semantics, monocular depth, and object poses in a single pass.

\item Experiments show that our model reconstructs real-world tabletop scenes with arbitrary numbers of objects, delivering more photorealistic and amodally complete results while running far faster than multi-stage pipelines.

\end{enumerate}
\section{Related Work}
\label{sec:related work}

\paragraph{Feed‑forward 3D reconstruction.} Feed‑forward 3D reconstruction maps one or a few images directly to a renderable 3D scene.
Previous works have explored 3D representations including voxel grids~\cite{tatarchenko2017octreegeneratingnetworksefficient, tulsiani2017multiviewsupervisionsingleviewreconstruction}, multi-plane images~\cite{li2021minecontinuousdepthmpi, tulsiani2018layerstructured3dsceneinference}, meshes~\cite{gkioxari2020meshrcnn, groueix2018atlasnetpapiermacheapproachlearning}, surfel~\cite{gao2023surfelnerfneuralsurfelradiance}, and radiance fields~\cite{hong2024lrmlargereconstructionmodel, yu2021pixelnerfneuralradiancefields}. More recently, 3D Gaussians~\cite{kerbl20233dgaussiansplattingrealtime} have emerged as a dominant representation for feed‑forward regression thanks to their real‑time differentiable rendering and compatibility with high‑capacity 2D backbones.

Early feed-forward Gaussian predictors focus on single, centered objects, assigning one Gaussian to each input pixel and directly regressing its parameters without test-time optimization~\cite{szymanowicz2024splatter, xu2024grmlargegaussianreconstruction, tang2024lgmlargemultiviewgaussian, zhang2024geolrmgeometryawarelargereconstruction}. These models deliver fast and high-quality reconstructions, but they assume clean, uncluttered inputs and cannot handle occlusions, multiple instances, or reassemble per-object predictions into a scene.

In contrast, scene-level Gaussian models predict a dense Gaussian field for an entire scene from one~\cite{szymanowicz2025flash3d} or multiple images~\cite{charatan2024pixelsplat3dgaussiansplats, zhang2024gslrmlargereconstructionmodel, ye2024poseproblemsurprisinglysimple, xu2025depthsplatconnectinggaussiansplatting, Chen_2024, wang2024freesplatgeneralizable3dgaussian}. Techniques including probabilistic splatting~\cite{charatan2024pixelsplat3dgaussiansplats}, cost-volume aggregation~\cite{Chen_2024}, depth conditioning~\cite{xu2025depthsplatconnectinggaussiansplatting}, pose-free formulations~\cite{ye2024poseproblemsurprisinglysimple} and large transformer backbones~\cite{zhang2024gslrmlargereconstructionmodel} have been explored to improve performance. While effective for novel-view synthesis, these scene-level approaches treat the world as a single undifferentiated cloud without instance decomposition, preventing downstream reasoning or interaction.

\paragraph{Object-centric scene reconstruction.} A separate line of work performs object-centric scene reconstruction, explicitly recovering a set of 3D object instances and their layout. IM2CAD~\cite{izadinia2017im2cad}, Total3DUnderstanding~\cite{nie2020total3dunderstandingjointlayoutobject}, Zhang \etal~\cite{zhang2021holistic3dsceneunderstanding}, and CoReNet~\cite{popov2020corenetcoherent3dscene} reconstruct indoor scenes from a single image by detecting furniture and room layout, then retrieving or predicting per-object geometry and enforcing consistency in a shared 3D frame. CAD- and RGB-D-based pipelines such as Mask2CAD~\cite{kuo2020mask2cad3dshapeprediction}, ROCA~\cite{Gumeli_2022_CVPR}, CenterSnap~\cite{irshad2022centersnapsingleshotmultiobject3d}, and ShAPO~\cite{irshad2022shapoimplicitrepresentationsmultiobject} further combine instance detection and depth with CAD retrieval or learned latent shape codes for each object, making them sensitive to upstream errors and computationally costly as the number of objects increases.

More recent methods introduce strong generative priors but largely retain this compositional, multi-stage design: Gen3DSR~\cite{Ardelean2025Gen3DSR}, CAST~\cite{yao2025castcomponentaligned3dscene}, and DepR~\cite{zhao2025deprdepthguidedsingleview} first apply monocular depth estimation and instance segmentation, then run object-level image-to-3D or diffusion models and compose the resulting objects into a coherent scene; MIDI~\cite{huang2025midimultiinstancediffusionsingle} extends pre-trained image-to-3D generators to a multi-instance diffusion model that still takes segmented object crops as input. Consequently, computational cost and brittleness scale with the number and quality of segmented instances. With a sufficiently large dataset and a sufficiently powerful model, we show that single-view, object-aware 3D reconstruction can be approached as a direct, feed-forward prediction problem, rather than a fragile sequence of segmentation, retrieval, optimization, or generative refinement. In practice, this shift yields reconstructions that are not only orders-of-magnitude faster but also higher-fidelity.

\section{Preliminaries}
\label{sec:preliminaries}
% In the following sections, we will use \textbf{bold} symbols to represent vectors, while \textit{plain} symbols to denote scalars and matrices.

\paragraph{3D Gaussian Splatting.} Gaussian Splatting~\cite{kerbl20233dgaussiansplattingrealtime} renders a scene represented by a finite set of anisotropic 3D Gaussian primitives by projecting each primitive to the image plane as a 2D Gaussian and alpha-compositing them in visibility order, yielding a fast, differentiable approximation to emission–absorption volume rendering. Compared with ray-sampled neural fields~\cite{mildenhall2020nerfrepresentingscenesneural}, splatting enables real-time rendering and efficient gradient backpropagation, and is widely used as the rendering backbone in recent feed-forward reconstruction methods~\cite{szymanowicz2024splatter, charatan2024pixelsplat3dgaussiansplats, Chen_2024, wang2024freesplatgeneralizable3dgaussian, ye2024poseproblemsurprisinglysimple, chen2024laraefficientlargebaselineradiance, tang2024lgmlargemultiviewgaussian, xu2024grmlargegaussianreconstruction, zhang2024gslrmlargereconstructionmodel, szymanowicz2025flash3dfeedforwardgeneralisable3d}; we adopt the same renderer throughout.

\paragraph{Normalized Object Coordinate Space.} Normalized Object Coordinate Space (NOCS) \cite{wang2019normalizedobjectcoordinatespace} assigns each 3D point on an object instance a category-level, pose-invariant coordinate $\mathbf c\in[0,1]^3$ within a unit canonical cube whose axes are consistently aligned across instances of that category. We denote this unit cube as the \emph{canonical space} and to its associated rigid coordinate system as the \emph{canonical frame}.

Dense per-pixel NOCS predictions $\hat{\mathbf c}_{u,v}$ provide pixel-to-canonical correspondences that, together with the predicted 3D point map, are sufficient to recover an instance's category-level pose $\Pi=(s,R,\mathbf t)\!\in\!\mathrm{SIM}(3)$ (\cref{subsec:inference}). By definition, $\Pi$ transforms canonical coordinates to the camera (or scene) frame: $\mathbf x^{\text{cam}} = \Pi(\mathbf x^{\text{can}}) = s R \mathbf x^{\text{can}} + \mathbf t$, with inverse mapping given by $\Pi^{-1}(\mathbf x^{\text{cam}})= s^{-1} R^\top\cdot (\mathbf x^{\text{cam}}-\mathbf t)$.

\section{Method}
\label{sec:method}

\begin{figure*}
    \centering
    \includegraphics[width=0.95\linewidth]{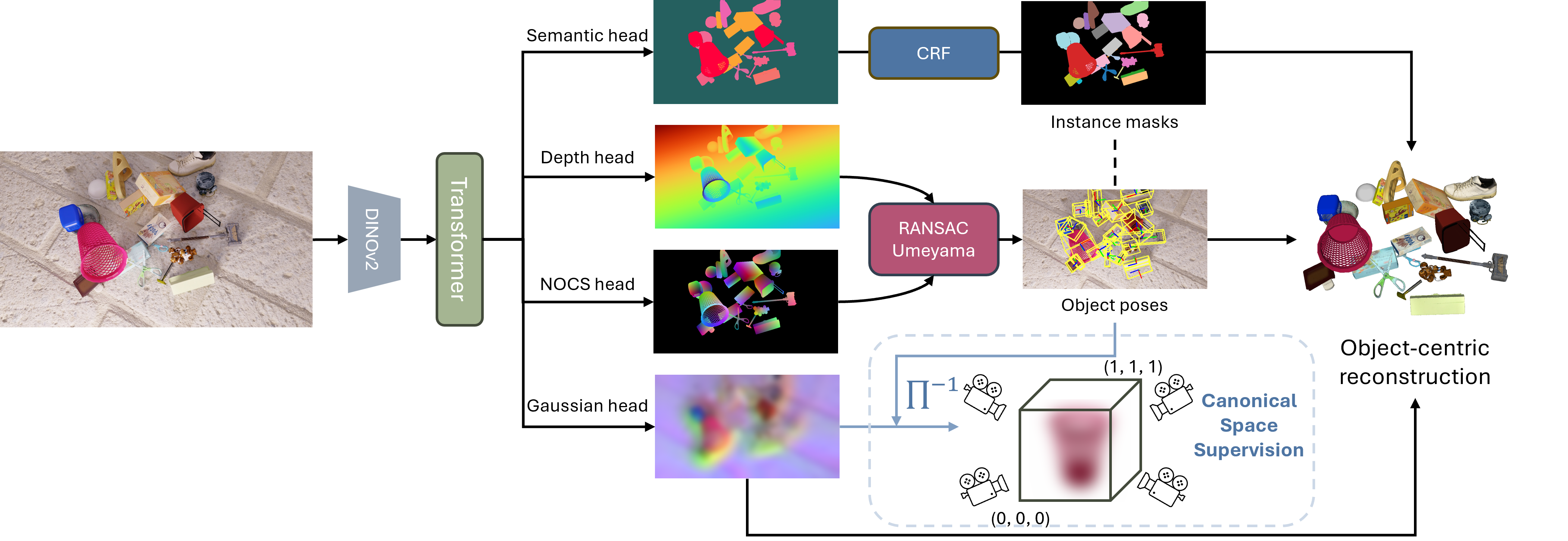}
    \vspace{-0.5em}
    \caption{\textbf{Overview of our single-view object-centric 3D reconstruction pipeline.
    }
Given a single RGB input, we extract dense DINOv2 features and feed them to a transformer that predicts per-pixel depth, CLIP-space semantic embeddings, NOCS coordinates, and Gaussian primitives. A CRF refines semantic affinities to produce coherent instance masks. For each instance, we estimate a category-level SIM(3) pose via RANSAC-Umeyama using the predicted NOCS-to-3D correspondences, enabling a transformation from camera space into the canonical object frame. The per-pixel Gaussians are then grouped and transformed into canonical space, where Canonical-Space Supervision (CSS) trains them to form amodally complete, compact 3D Gaussians. Aggregating all reconstructed objects yields an interactive, object-aligned scene representation from a single image.}
    \label{fig:pipeline}
    \vspace{-1em}
\end{figure*}

\subsection{Problem formulation and notation}
\label{subsec:notation}

Given a single RGB image $I \in \mathbb{R}^{H \times W \times 3}$ with unknown intrinsics $K$, \method converts the image into an object‑centric scene: a set of instances, each with a category‑level $\mathrm{SIM}(3)$ pose and a high-fidelity 3D representation. Achieving this in one pass requires pixel‑aligned predictions that are sufficient to (i) discover instances and semantics, (ii) recover a metric similarity transform for each object, and (iii) assemble amodally complete Gaussian representation of each object into an interactive scene.

Specifically, for each pixel $(u, v)$, our network $\Phi$ outputs: \begin{equation}\Phi(I)_{u, v} = (\hat{\mathbf{e}}_{u, v}, \hat{d}_{u, v}, \hat{\mathbf{c}}_{u, v}, \hat{\mathcal{G}}_{u, v}),\end{equation} where $\hat{\mathbf{e}}_{u, v} \in \mathbb{R}^{512}$ is the semantic label of the object that this pixel belongs to. We define semantic label of an object as the CLIP~\cite{radford2021learningtransferablevisualmodels} embedding of the object's category name~\cite{li2022languagedrivensemanticsegmentation}. $\hat{d}_{u, v} \in \mathbb{R}^+$ is the predicted metric depth. It enables back-projecting the pixel into 3D space via $\hat{\mathbf{p}}_{u, v} = \hat{d}_{u, v} \cdot K^{-1} \begin{bmatrix}u & v&1\end{bmatrix}^{\top} \in \mathbb{R}^3$. $\hat{\mathbf{c}}_{u, v} \in [0,1]^3$ is the predicted NOCS~\cite{wang2019normalizedobjectcoordinatespace} coordinate, which enables $\mathrm{SIM}(3)$ pose recovery.

$\hat{\mathcal{G}}_{u,v} = \{g_{u,v}^{(i)}\}_{i = 1}^{k}$ is a small set of anisotropic 3D Gaussian primitives (we use $k = 2$) that will be aggregated into per‑object reconstruction: \begin{equation}g_{u,v}^{(i)} = (\boldsymbol{\mu}_{u,v}^{(i)}, \Sigma_{u, v}^{(i)}, \alpha_{u,v}^{(i)}, \mathbf{S}_{u,v}^{(i)}),\end{equation} with mean $\boldsymbol{\mu}_{u,v}^{(i)} \in \mathbb{R}^3$, covariance $\Sigma_{u, v}^{(i)}\in \mathbb{S}^3_{++}$, opacity $\alpha_{u,v}^{(i)} \in (0, 1)$, and RGB spherical harmonics (SH) coefficients $\mathbf{S}_{u,v}^{(i)} \in \mathbb{R}^{3(L + 1)^2}$ (order $L$). The Gaussian parameters are defined and predicted in camera frame, and will be transformed into each object's canonical frame for supervision and inference (Sec.~\ref{subsec:css}).

Following VGGT~\cite{wang2025vggtvisualgeometrygrounded}, we also predict the camera field of view $(\hat{\theta}_w,\hat{\theta}_h)$ of the input image and construct $K$ with $f_w=\frac{W}{2\tan(\hat{\theta}_w/2)}$, $f_h=\frac{H}{2\tan(\hat{\theta}_h/2)}$ and principal point at image center.

% \paragraph{Why these predictions?} 

In \cref{subsec:inference}, we show how $\hat{\mathbf e}$, $\hat d$, $\hat{\mathbf{c}}$, and $\hat{\mathcal G}$ are used to discover instances, estimate object poses, and assemble reconstructed objects. \cref{subsec:css} introduces Canonical Space Supervision (CSS) that trains Gaussians to be object‑aligned and amodally complete without per‑image Gaussian labels. \cref{subsec:architecture} summarizes architectural and training details. Our full pipeline is given in \cref{fig:pipeline}.

\subsection{Assembling objects from dense predictions}
\label{subsec:inference}

% At test time we treat the dense predictions as an oracle and perform a lightweight, deterministic assembly.

\paragraph{Instance discovery.} 
At inference time, for each pixel, we first compute the cosine similarity between the predicted embedding $\hat{\mathbf{e}}_{u,v}$ and a set of predefined category name CLIP embeddings $\{\mathbf{l}_c\}$. We then apply a fully connected conditional random field (CRF)~\cite{krahenbuhl2011efficient}, using unary potentials defined as $-\log\frac{\exp(\cos(\hat{\mathbf{e}}_{u,v}, \mathbf{l}_c) / \tau)}{\sum_{c'} \exp(\cos(\hat{\mathbf{e}}_{u,v}, \mathbf{l}_{c'}) / \tau)}$ for each category $c$, where $\tau$ denotes the temperature parameter in the softmax function. Pairwise potentials are defined as $\cos(\hat{\mathbf{e}}_{u,v},\hat{\mathbf{e}}_{u',v'})$. For more details about CRF, we refer the reader to \cite{krahenbuhl2011efficient}. This process yields groups $\{\hat{\mathcal{P}}_j\}$, where each $\hat{\mathcal{P}}_j$ represents the set of pixels corresponding to object $j$.

\paragraph{Pose estimation.} With the pixels for each object instance identified, we use their predicted NOCS~\cite{wang2019normalized} coordinates $\mathbf{c}_{u,v}$ to determine the object's precise $\mathrm{SIM}(3)$ pose in the scene. The NOCS coordinates establish a correspondence between a point's observed position in the scene and its standardized position within a unit canonical cube. We can therefore solve for the similarity transformation $\hat{\Pi}_j = (\hat{s}_j, \hat{R}_j, \hat{\mathbf{t}}_j)$, representing the scale, rotation, and translation, which maps the canonical space of object $j$ to the camera space. This is achieved by minimizing the alignment error between the back-projected 3D points and the transformed NOCS coordinates over all pixels belonging to that instance: \begin{equation}\hat{\Pi}_j = \arg\min_{\Pi} \sum_{(u,v) \in \hat{\mathcal{P}}_j} \left\| \hat{\mathbf{p}}_{u,v} - \Pi(\hat{\mathbf{c}}_{u,v}) \right\|^2,\end{equation} where $\Pi(\hat{\mathbf{c}}_{u,v}) = sR\cdot\hat{\mathbf{c}}_{u,v} + \mathbf{t}$. This optimization can be solved using Umeyama algorithm~\cite{88573} with RANSAC~\cite{10.1145/358669.358692}. The resulting inverse transformation, $\hat{\Pi}_j^{-1}$, gives us a direct mapping from the cluttered scene into the clean canonical space for each object. Notably, this NOCS prediction can also be used to differentiate object instances with the same category name but that are adjacent in the mask, where CRF alone may not be enough. We run multiple RANSACs within each CRF-generated mask, and output objects when there are still enough inliers.

\paragraph{Object Gaussians. } With the instance mask and estimated pose in hand, we obtain each object's canonical-space Gaussian representation by transforming every predicted Gaussian mean as $\boldsymbol{\mu}^{\text{can}, (i)}_{u,v} = \hat{\Pi}_j^{-1}(\boldsymbol{\mu}^{(i)}_{u,v})$ for all $(u, v) \in \hat{\mathcal{P}}_j, i \in \{1, \dots, k\}$. The resulting set of transformed Gaussians forms the complete canonical representation of object $j$.

\paragraph{Efficiency.}
% As proven in Appendix xxx, \method's inference runs in $\Theta(HW)$ time, 
Since \method predicts all per-pixel quantities in one forward pass, every object is reconstructed at once. Our CUDA CRF runs in roughly 200 ms per image, and our CUDA RANSAC Umeyama adds under 10 ms per object, making its cost negligible. Consequently, runtime is nearly invariant to scene complexity and remains far below prior pipelines~\cite{yao2025castcomponentaligned3dscene, tang2024diffuscenedenoisingdiffusionmodels, zhao2025deprdepthguidedsingleview}, which synthesize each object through iterative diffusion denoising and often require relation-graph optimization that grows quadratically with the number of objects.

\subsection{Canonical‑Space Supervision}
\label{subsec:css}

\begin{figure}
    \centering
    \includegraphics[width=\linewidth]{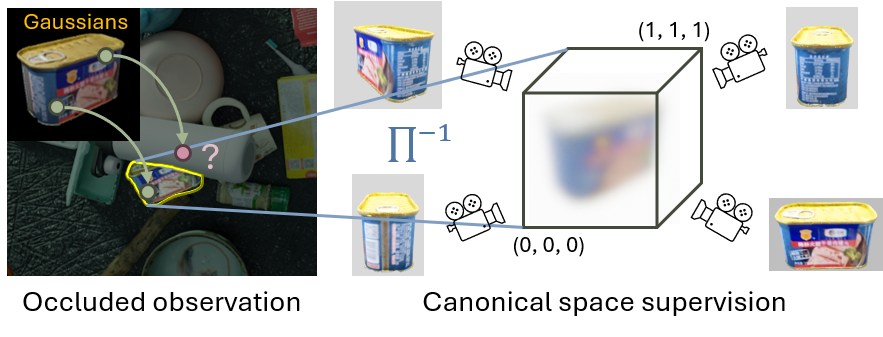}
    \caption{
\textbf{Canonical Space Supervision (CSS).}
Predicted per-pixel Gaussians are transformed into the object’s canonical frame via the ground-truth pose $\Pi^{-1}$. 
In canonical space, they are supervised against pre-rendered multi-view ground-truth images, providing clean amodal signals that resolve occlusions and enforce compact, object-aligned Gaussian reconstructions.
}\vspace{-1em}
    \label{fig:css}
\end{figure}

One key challenge for our Gaussian prediction network is that it must infer a full, amodal set of object Gaussians from only the pixels that are actually visible. A natural idea is to use pre-optimized object Gaussians and place them in the camera frame so they can serve as ground-truth supervision. However, there lacks one-to-one correspondence between visible pixels and ground-truth object Gaussians. To address this, we introduce \textit{Canonical Space Supervision} (\cref{fig:css}), a strategy that transfers training signals to the object's canonical frame, where clean targets are available.

Concretely, we place each training object mesh in the canonical frame, and pre‑render a set of $N$ views $\mathcal{V} = \{I_{n}^{\text{gt}}\}_{n = 1}^{N}$, where $I_n^{\text{gt}} \in \mathbb{R}^{H_{\text{can}} \times W_{\text{can}} \times 3}$. We set $N=42$, $H_{\text{can}} = W_{\text{can}} = 512$. This is done once per object and reused across all images containing that object.

During training, we use ground truth object masks $\{\mathcal{P}_j\}$ to extract pixels of each object $j$. 
% For each pixel $(u, v) \in \mathcal{P}_j$ and $i \in \{1,\dots,k\}$, 
We transform the predicted Gaussians per object from the camara space into the canonical space with the ground truth object pose.

With the transformed Gaussians $\{g_{u, v}^{\text{can}, (i)}\}$ of the object, we can render images of that object in its canonical space using the same camera angles as $\mathcal{V}$ with a differentiable Gaussian rasterizer~\cite{kerbl20233d}. Let $\{\hat{I}_{n}\}_{n = 1}^N$ denote the rendered images. Then the CSS loss is calculated by \begin{equation}
    \mathcal{L}_{\text{CSS}} = \sum_{n = 1}^{N} \Big(\|I_{n}^{\text{gt}} - \hat{I}_n\|_1 + \lambda_{\text{SSIM}} (1 - \mathrm{SSIM}(I_n^{\text{gt}} , \hat{I}_n))\Big).
\end{equation}

\paragraph{Occlusion handling via off‑ray offsets.} Each Gaussian mean is anchored at $\hat{\mathbf p}_{u, v}$ with a predicted camera‑space offset $\boldsymbol{\Delta}^{(i)}_{u, v}$, allowing a visible pixel to spawn Gaussians behind the first surface. Because CSS supervises in canonical space, occluded Gaussians receive gradients even when not visible in the input. Optionally, we regularize with an annealed small‑offset prior: $\mathcal{L}_{\text{reg}} = \sum_{(u, v)\in \mathcal{P}_j}\sum_{i = 1}^{k} \mathrm{ReLU}(\|\boldsymbol{\Delta}^{(i)}_{u, v} - \tau_{\text{offset}}\|_1)$.

\subsection{Architecture and training details}
\label{subsec:architecture}

\paragraph{Architecture.}
Inspired by VGGT~\cite{wang2025vggtvisualgeometrygrounded}, \method uses a DINOv2 backbone~\cite{oquab2024dinov2learningrobustvisual} followed by a 48‑layer ViT encoder and DPT‑style decoder heads~\cite{ranftl2021visiontransformersdenseprediction}.
The input image is patchified by DINOv2 and processed by global self‑attention through the encoder.

For dense tasks, each head takes features from four intermediate encoder layers (lateral skips), projects them to a common width, and upsamples to the image resolution with convolutional fusion.

For camera FOV prediction, we append a learnable camera token that is updated by a small stack of transformer blocks with adaptive layer‑norm modulation and iterative refinement; a linear layer regresses the field‑of‑view angles $(\hat\theta_w,\hat\theta_h)$.%
\footnote{Our implementation retains VGGT’s iterative refinement; translation/rotation channels are present in the token state but only FOV is used at test time.}

For Gaussian prediction we decouple geometry and appearance.
A geometry head outputs per‑pixel off‑ray offsets $\boldsymbol{\Delta}^{(i)}_{u,v}$ (for $i{=}1,\dots,k$), which are added to the back‑projected point $\hat{\mathbf p}_{u,v}$ to obtain camera‑frame means $\boldsymbol{\mu}^{(i)}_{u,v}$.
An appearance/shape head predicts canonical‑frame scales $\boldsymbol{\sigma}^{(i)}_{u,v}$, unit quaternions $\mathbf q^{\mathrm{can},(i)}_{u,v}$, opacities $\alpha^{(i)}_{u,v}$, and SH coefficients $\mathbf S^{(i)}_{u,v}$.
Following NoPoSplat~\cite{ye2024poseproblemsurprisinglysimple}, we provide an \emph{RGB shortcut} to the appearance/shape head to improve fine texture details in 3D reconstruction.
All parameters of each Gaussian and the $k$ Gaussians of every pixel are simply concatenated.

\paragraph{Training.}
We optimize all tasks jointly with AdamW and a cosine learning‑rate schedule. DINOv2 is initialized from public weights and fine‑tuned end‑to‑end; the full list of hyperparameters is provided in the appendix.

\emph{Depth.}
We supervise canonical inverse depth as in Depth Pro~\cite{bochkovskii2025depthprosharpmonocular}. Let $f_{w}$ be the horizontal focal length (pixels) and $W$ the image width, define $C = \frac{f_{w}}{W\cdot d}$. Our model outputs $\hat{C}$ and is trained to minimize $\mathcal{L}_{\text{depth}}
= \|\hat C - C\|_2
+ \lambda_\nabla\!\left(\|\nabla_x(\hat C - C)\|_2 + \|\nabla_y(\hat C - C)\|_2\right)$. At test time, we recover metric depth by $\hat d_{u,v}\coloneqq\frac{\hat f_{\mathrm{px}}}{W\cdot \hat C_{u,v}}$.

\emph{Semantics.}
% We manually curate a vocabulary of 233 everyday labels that cover all the training objects and precompute their CLIP text embeddings. 
During training time, we dynamically compute cosine similarities between the embedding of pixel and all the words that appear in the training image. We then encourage the embedding of the pixel to align with the ground-truth class by using the computed cosine similarities as the logits for $\mathrm{softmax}$, with cross entropy loss.

\emph{NOCS.}
We reformulate NOCS coordinate regression as a bin classification task augmented with a learnable offset, which implicitly resolves ambiguities in symmetric objects. Each axis in NOCS (i.e., \textit{xyz}) is discretized into $M{=}64$ centered bins. We supervise the bin classification using a cross-entropy loss and the offset prediction using a mean squared error loss. The total loss is averaged over all foreground object pixels.

\emph{Gaussians (CSS).}
Gaussian supervision follows Canonical Space Supervision discussed in Sec.~\ref{subsec:css}.

\emph{Camera FOV.} We supervise $(\hat\theta_w,\hat\theta_h)$ with a robust Huber loss on angles:
$\mathcal L_{\mathrm{cam}}=\|\hat\theta_w-\theta_w\|_{\epsilon}+\|\hat\theta_h-\theta_h\|_{\epsilon}$.

Task losses are combined with homoscedastic uncertainty weighting~\cite{kendall2018multitasklearningusinguncertainty}; ablations are in the appendix.
\section{Dataset}
\label{sec:dataset}
% \paragraph{Datasets.}
Existing indoor benchmarks for monocular depth estimation and open vocabulary semantic segmentation \cite{SilbermanECCV12, Song:2015:SRA, dai2017scannetrichlyannotated3dreconstructions} primarily emphasize room layout and large furniture, while providing limited coverage of the object interaction scale that is central to everyday visual tasks. In contrast, progress in embodied perception \cite{wen2024foundationposeunified6dpose, shridhar2022perceiveractormultitasktransformerrobotic, brohan2023rt2visionlanguageactionmodelstransfer, Fang_2020_CVPR} and in mobile AR or MR systems \cite{10.1145/3379337.3415881, Holynski2018Occlusion} requires accurate modeling of small, manipulable, and semantically diverse objects such as cups, tools, and containers that humans routinely interact with.

To support this direction, we construct a new evaluation benchmark by integrating several real world datasets tailored to this domain. Specifically, we include the validation split of HOPE \cite{tyree20226dofposeestimationhousehold} and the test splits of YCB Video \cite{xiang2018posecnn}, PACE \cite{you2024pacelargescaledatasetpose}, Omni6DPose \cite{zhang2024omni6dposebenchmarkmodeluniversal} (OMNI), and NOCS \cite{wang2019normalizedobjectcoordinatespace}. For training, we curate and align four large scale sources: the training splits of PACE and Omni6DPose, Google Scanned Objects \cite{downs2022googlescannedobjectshighquality} renderings from FoundationPose \cite{wen2024foundationposeunified6dpose}, and HyperSim \cite{roberts2021hypersimphotorealisticsyntheticdataset}.

\section{Experiments}

We first evaluate holistic 3D object-centric reconstruction from a single RGB image in \cref{sec:rec}. \cref{sec:zeroshot} then demonstrates that beyond 3D reconstruction, our method also delivers state-of-the-art zero-shot performance on depth estimation, segmentation, and object pose prediction. \cref{sec:ablation} examines key design choices through targeted ablations.

\subsection{3D Reconstruction}
\label{sec:rec}

We evaluate \method against Gen3DSR \cite{Ardelean2025Gen3DSR}, ACDC \cite{dai2024acdc}, and a unified glued baseline that uses instance masks from SAM2 \cite{ravi2024sam} and GroundingDINO \cite{liu2024grounding}, object poses from MonoDiff9D \cite{MonoDiff9D}, and depth from DepthPro \cite{bochkovskii2025depthprosharpmonocular}. We denote this pipeline as AoE (Army of Experts). Since the baselines are extremely slow, we randomly sample ten images from each dataset in our benchmark.

Following prior work \cite{Ardelean2025Gen3DSR, yao2025castcomponentaligned3dscene}, we report Chamfer Distance and F-1@0.1 between the predicted and ground truth meshes, and CLIP similarity between the rendered and ground truth images. All backgrounds are manually normalized to white for both predictions and ground truth.

\begin{figure*}
    \centering
    \includegraphics[width=0.9\linewidth]{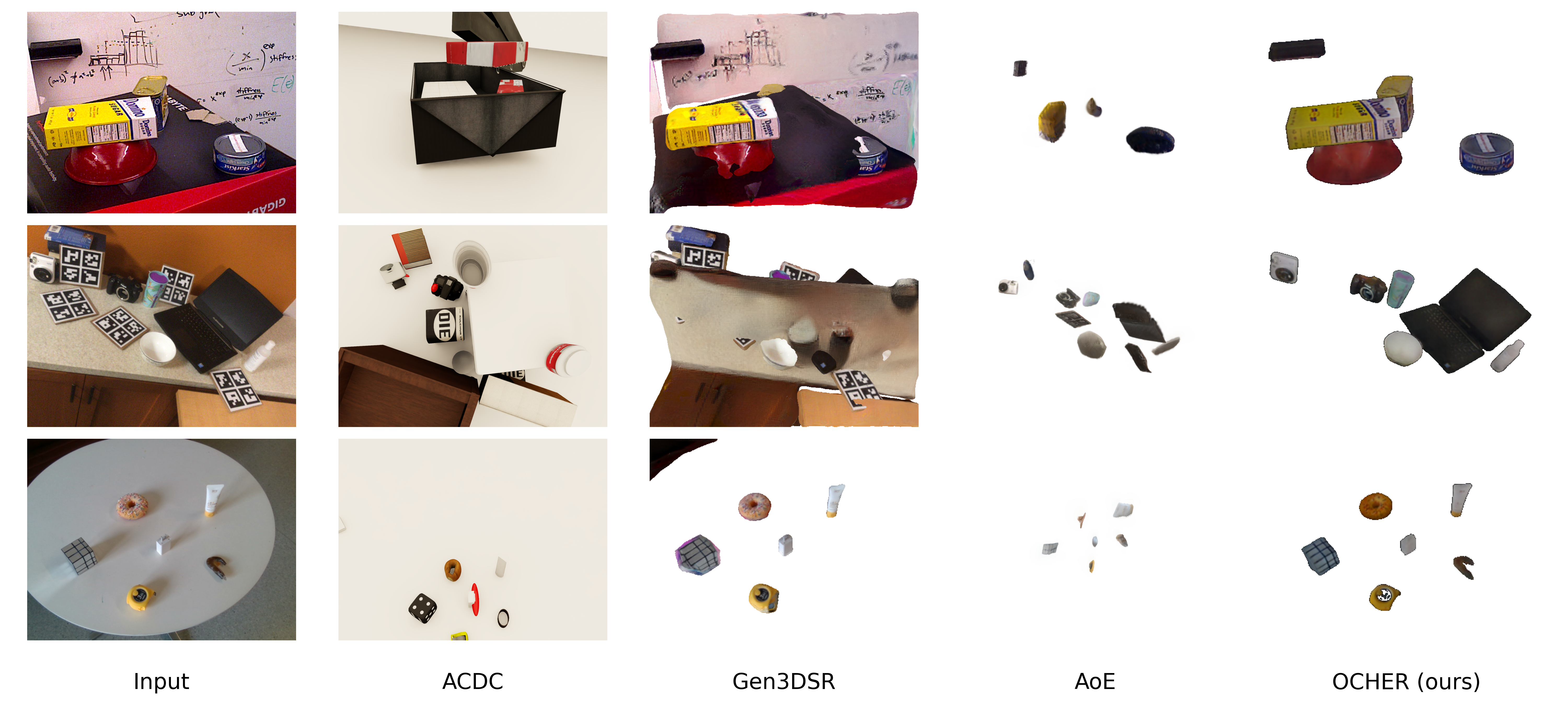}
    \vspace{-1em}
    \caption{
\textbf{Qualitative comparison of single-image 3D object-centric reconstruction.}
Given a single RGB input, we compare our method (\method) with ACDC, Gen3DSR, and AoE (Army of Experts: SAM2 + GroundingDINO + MonoDiff9D + DepthPro). 
Prior methods often yield incomplete geometry, distorted textures, or missing objects. 
\method reconstructs sharper, more complete, and semantically consistent objects across diverse scenes.
}
    \label{fig:rec}
\end{figure*}

As shown in \cref{tab:cd_f1_clip}, our method establishes a clear margin over all baselines. On PACE, \method reduces the Chamfer Distance from 0.31 (Gen3DSR) and 0.35 (AoE) to 0.18, while more than doubling the best F-1 score (45.00 versus AoE's 21.39). Similar trends hold across all remaining datasets: on YCB-V, \method achieves 0.17 CD (a 26 percent improvement over AoE and a 48 percent improvement over Gen3DSR) and reaches 22.71 F-1, surpassing the strongest baseline by over 10 points. On HOPE, \method attains 83.69 CLIP similarity, exceeding Gen3DSR by +6.1 and AoE by +19.9. For NOCS real, \method's gains are the most pronounced, improving CD from 0.15 to 0.07 and F-1 from 38.01 to 76.77. These accuracy improvements come alongside a dramatic speedup: our 0.7\,s inference time per image is roughly 2,000x faster than Gen3DSR (25.6\,min) and ACDC (22.1\,min), while also running more than 30x faster than AoE (21.6\,s). It demonstrates the advantage of our unified per pixel prediction formulation. Some qualitative results are shown in \cref{fig:rec}.

\begin{table*}[ht]
\centering
\small
\setlength{\tabcolsep}{4pt}
\resizebox{\textwidth}{!}{%
\begin{tabular}{lcccccccccccccccc}
\toprule
\multirow{2}{*}{Model} &
\multicolumn{3}{c}{PACE} & 
\multicolumn{3}{c}{OMNI} &
\multicolumn{3}{c}{YCB-V} &
\multicolumn{3}{c}{HOPE} &
\multicolumn{3}{c}{NOCS real} & \multirow{2}{*}{Time per image$\downarrow$} \\
\cmidrule(lr){2-4} \cmidrule(lr){5-7} \cmidrule(lr){8-10} \cmidrule(lr){11-13} \cmidrule(lr){14-16}
 & CD$\downarrow$ & F-1$\uparrow$ & CLIP$\uparrow$ 
 & CD$\downarrow$ & F-1$\uparrow$ & CLIP$\uparrow$ 
 & CD$\downarrow$ & F-1$\uparrow$ & CLIP$\uparrow$ 
 & CD$\downarrow$ & F-1$\uparrow$ & CLIP$\uparrow$ 
 & CD$\downarrow$ & F-1$\uparrow$ & CLIP$\uparrow$ & \\
\midrule

ACDC~\cite{dai2024acdc} & 0.69 & 0.00 & 73.74 & -- & -- & 72.23 & 0.32 & 7.76 & 60.15 & 1.42 & 1.92 & 63.75 & 3.44 & 1.72 & 69.74 & 22.1~min\\
Gen3DSR~\cite{Ardelean2025Gen3DSR} & \underline{0.31} & 12.93 & \underline{81.61} & -- & -- & \underline{82.00} & 0.33 & 8.85 & \underline{75.26} & 0.35 & 12.05 & \underline{77.63} & 0.26 & 17.73 & \underline{75.33} & 25.6~min\\
AoE & 0.35 & \underline{21.39} & 77.97 & -- & -- & 69.31 & \underline{0.23} & \underline{12.45} & 66.05 & \underline{0.18} & \textbf{47.71} & 63.80 & \underline{0.15} & \underline{38.01} & 70.62 & 21.6~s\\
\textbf{\method (ours)} & \textbf{0.18} & \textbf{45.00} & \textbf{83.15} & -- & -- & \textbf{82.78} & \textbf{0.17} & \textbf{22.71} & \textbf{78.80} & \textbf{0.14} & \underline{40.08} & \textbf{83.69} & \textbf{0.07} & \textbf{76.77} & \textbf{85.90} & \textbf{0.7}~\textbf{s}\\

\bottomrule
\end{tabular}
}
\caption{Comparison of 3D reconstruction and semantic consistency across datasets using CD (Chamfer Distance, lower is better), F-1 score, and CLIP similarity.}
\vspace{-1em}
\label{tab:cd_f1_clip}
\end{table*}

\begin{table*}[t]
\centering
\small
\setlength{\tabcolsep}{4pt}
\resizebox{\textwidth}{!}{%
\begin{tabular}{lccccccccccccccc}
\toprule
\multirow{2}{*}{Model} & \multicolumn{3}{c}{PACE} & \multicolumn{3}{c}{OMNI} & \multicolumn{3}{c}{YCB-V} & \multicolumn{3}{c}{HOPE} & \multicolumn{3}{c}{NOCS~real} \\
\cmidrule(lr){2-4} \cmidrule(lr){5-7} \cmidrule(lr){8-10} \cmidrule(lr){11-13} \cmidrule(lr){14-16}
 & $\delta_1$$\uparrow$ & AbsRel$\downarrow$ & RMSE$\downarrow$ & $\delta_1$$\uparrow$ & AbsRel$\downarrow$ & RMSE$\downarrow$ & $\delta_1$$\uparrow$ & AbsRel$\downarrow$ & RMSE$\downarrow$ & $\delta_1$$\uparrow$ & AbsRel$\downarrow$ & RMSE$\downarrow$ & $\delta_1$$\uparrow$ & AbsRel$\downarrow$ & RMSE$\downarrow$ \\
\midrule
NeWCRFs~\cite{yuan2022newcrfsneuralwindow} & 17.00 & 0.7356 & 0.5308 & 21.41 & 0.6432 & 0.9934 & 52.71 & 0.2740 & 0.2699 & 5.37 & 0.8073 & 0.5029 & 36.00 & 0.3352 & 0.3586 \\

ZoeDepth~\cite{bhat2023zoedepthzeroshottransfercombining} & 8.20 & 0.8508 & 0.6146 & 20.73 & 0.5794 & 0.9406 & 52.84 & 0.2704 & 0.2750 & 0.53 & 0.9814 & 0.6055 & 34.08 & 0.4399 & 0.5275 \\

Metric3D V2~\cite{Hu_2024} & 1.32 & 0.4695 & 0.3995 & \textbf{73.73} & \textbf{0.1788} & \underline{0.7963} & 15.44 & 0.2791 & 0.2538 & 0.01 & 0.4039 & 0.2968 & 8.50 & 0.3273 & 0.3571 \\

Depth Anything V2~\cite{yang2024depthv2} & 13.07 & 0.7373 & 0.5906 & 5.83 & 0.9009 & 1.1728 & 50.74 & \underline{0.1899} & \textbf{0.2029} & 0.07 & 1.0364 & 0.6518 & 2.44 & 0.5507 & 0.6088 \\

Depth Pro~\cite{bochkovskii2025depthprosharpmonocular} & 33.63 & 0.5603 & 0.5610 & \underline{57.37} & 0.2742 & \textbf{0.7940} & 49.04 & 0.2078 & 0.2148 & 23.09 & \underline{0.4035} & 0.3013 & 81.58 & 0.1370 & 0.1831 \\

VGGT~\cite{wang2025vggtvisualgeometrygrounded} & \underline{55.58} & \underline{0.2415} & \underline{0.2096} & 57.14 & \underline{0.2272} & 1.0365 & \underline{63.88} & \textbf{0.1822} & 0.2131 & \underline{37.17} & 0.4313 & \underline{0.2844} & \underline{92.26} & \underline{0.0965} & \underline{0.1394} \\

UniDepth V2~\cite{piccinelli2025unidepthv2universalmonocularmetric}                         & 35.05 & 0.5076 & 0.4820 & 22.80 & 0.7221 & 1.1007 & 32.70 & 0.2548 & 0.2499 & 8.63 & 0.7622 & 0.6830 & 78.92 & 0.1778 & 0.2366 \\
\midrule
\textbf{\method (ours)}                        & \textbf{94.82} & \textbf{0.0834} & \textbf{0.1039} & 42.77 & 0.2900 & 0.8042 & \textbf{69.96} & 0.1933 & \underline{0.2044} & \textbf{61.36} & \textbf{0.2192} & \textbf{0.1812} & \textbf{98.43} & \textbf{0.0923} & \textbf{0.1066} \\
\bottomrule
\end{tabular}
}
\caption{\textbf{Monocular metric depth estimation results} on PACE, OMNI, YCB-V, HOPE, and NOCS~real. Each dataset block reports $\delta_1$ (in percentage), AbsRel, and RMSE. Bold indicates the best result, and underline indicates the second best.}
\label{tab:depth-mono}
\end{table*}

\begin{table*}[t]
\centering
\small
\setlength{\tabcolsep}{4pt}
\resizebox{\textwidth}{!}{%
\begin{tabular}{lccccccccccccccc}
\toprule
\multirow{2}{*}{Model} & \multicolumn{3}{c}{PACE} & \multicolumn{3}{c}{OMNI} & \multicolumn{3}{c}{YCB-V} & \multicolumn{3}{c}{HOPE} & \multicolumn{3}{c}{NOCS~real} \\
\cmidrule(lr){2-4} \cmidrule(lr){5-7} \cmidrule(lr){8-10} \cmidrule(lr){11-13} \cmidrule(lr){14-16}
 & mIoU$\uparrow$ & $\mathrm{FB\text{-}IoU}$$\uparrow$ & hit@5$\uparrow$ & mIoU$\uparrow$ & $\mathrm{FB\text{-}IoU}$$\uparrow$ & hit@5$\uparrow$ & mIoU$\uparrow$ & $\mathrm{FB\text{-}IoU}$$\uparrow$ & hit@5$\uparrow$ & mIoU$\uparrow$ & $\mathrm{FB\text{-}IoU}$$\uparrow$ & hit@5$\uparrow$ & mIoU$\uparrow$ & $\mathrm{FB\text{-}IoU}$$\uparrow$ & hit@5$\uparrow$ \\
\midrule
LSeg~\cite{li2022languagedrivensemanticsegmentation} & 0.26 & 17.23 &  8.52 &  2.43 &  4.55 & 16.31 & 2.63 & 12.41 & 23.05 & 0.30 & 12.09 & 20.69 &  5.48 &  8.73 & 79.82 \\
OVSeg~\cite{liang2023openvocabularysemanticsegmentationmaskadapted} & 1.28 & 13.57 & 28.84 & 11.24 &  2.94 & 52.19 & 1.29 & 12.20 & 58.66 & 0.31 & 11.87 & 25.99 &  1.87 &  8.54 & 79.21 \\
ODISE~\cite{xu2023openvocabularypanopticsegmentationtexttoimage} & 1.69 & 13.70 & 36.05 & 13.27 &  3.66 & 58.54 & 4.51 & 12.17 & 76.76 & 0.99 & 11.87 & 68.77 &  6.31 &  8.53 & 90.60 \\
FC-CLIP~\cite{yu2023convolutionsdiehardopenvocabulary}         & \underline{3.67} & 13.56 & 61.62 & \underline{22.29} &  3.61 & 71.11 & 3.69 & 12.15 & 69.96 & 1.85 & 11.91 & 71.80 &  4.72 &  9.40 & 94.68 \\
CAT-Seg~\cite{cho2024catsegcostaggregationopenvocabulary}         & 1.26 & 21.48 & 30.53 &  9.61 &  4.14 & 52.62 & 2.38 & 12.12 & 65.78 & 2.87 & 11.91 & 75.97 &  5.31 &  8.79 & 95.80 \\
SAN~\cite{xu2023adapternetworkopenvocabularysemantic}             & 1.70 & \underline{71.05} & 31.87 & 13.75 & \underline{40.82} & 53.09 & 4.25 & \underline{44.77} & 72.83 & \underline{3.27} & \underline{29.96} & 63.94 &  5.22 & \underline{45.30} & 94.56 \\
MAFT+~\cite{jiao2024collaborativevisiontextrepresentationoptimizing}           & 3.66 & 13.76 & \underline{64.39} & \textbf{23.02} &  3.61 & \textbf{76.69} & \underline{5.00} & 12.16 & \underline{79.01} & 3.21 & 11.87 & \textbf{89.23} & \underline{8.10} &  8.53 & \underline{96.14} \\
\midrule
\textbf{\method (ours)}    & \textbf{7.34} & \textbf{94.95} & \textbf{84.76} & 18.43 & \textbf{84.21} & \underline{75.23} & \textbf{9.14} & \textbf{72.00} & \textbf{80.03} & \textbf{6.90} & \textbf{71.78} & \underline{79.27} & \textbf{13.40} & \textbf{92.95} & \textbf{98.51} \\
\bottomrule
\end{tabular}
}
\caption{\textbf{Open-vocabulary semantic segmentation results} on PACE, OMNI, YCB-V, HOPE, and NOCS~real. Each dataset block reports mIoU, FB-IoU, and hit@5 in percentages. Bold indicates the best result, and underline indicates the second best.}
\label{tab:ovss-full}
\vspace{-0.5em}
\end{table*}

\subsection{Individual task performance}
\label{sec:zeroshot}

\paragraph{Zero-shot metric depth.} Accurate metric depth from a single RGB image is essential to our pipeline, as it defines the anchor positions for our 3D Gaussians. We evaluate \method on five datasets against seven state-of-the-art baselines using three standard metrics: $\delta_1$~\cite{Ladicky_2014_CVPR}, AbsRel, and RMSE. Additional metrics ($\delta_2$, $\delta_3$, log10, $\mathrm{RMSE}_{\mathrm{log}}$, SI-log) are provided in the Supplementary.

As shown in \cref{tab:depth-mono}, \method achieves leading performance on all three metrics for PACE, HOPE, and NOCS-real, and on $\delta_1$ for YCB-V. Although Metric3D V2~\cite{Hu_2024} and Depth Pro~\cite{bochkovskii2025depthprosharpmonocular} perform slightly better on OMNI, the margin is minimal; moreover, Metric3D V2 requires ground-truth camera intrinsics, giving it an inherent advantage, yet it is still surpassed by \method on four of the five datasets. Depth Anything V2~\cite{yang2024depthv2}, trained primarily for relative depth and fine-tuned for metric estimation, shows high domain sensitivity, excelling on YCB-V (narrowly ahead of \method) but degrading substantially elsewhere. Overall, \method attains the best results in 10 of 15 metric–dataset combinations and delivers the strongest average performance across benchmarks.

\paragraph{Zero-shot semantic segmentation.}
Open-vocabulary semantic segmentation assigns per-pixel labels drawn from a potentially open set of natural-language concepts. To build an evaluation vocabulary disjoint from training, we aggregate names from the test datasets, common indoor categories from ADE20K~\cite{zhou2017scene, zhou2019semantic}, and additional household items.

We report standard OVSS metrics: mIoU and FB-IoU. Given the difficulty of segmenting fine-grained, cluttered indoor scenes, we additionally report hit@5, which allows each method to produce up to five candidate labels per pixel. Since FB-IoU already captures the ability to separate foreground from background, the remaining metrics are computed on foreground regions only.

\cref{tab:ovss-full} compares \method with seven OVSS baselines. Across five datasets and three metrics (15 settings), \method ranks first in 12 and achieves the best average rank (1.27). It leads all three metrics on PACE, YCB-V, and NOCS-Real; mIoU and FB-IoU on HOPE; and FB-IoU on OMNI. Averaged across datasets, \method obtains 11.04 mIoU, 83.18 FB-IoU, and 83.56 hit@5, outperforming the strongest baseline by +2.44 mIoU (MAFT+~\cite{jiao2024collaborativevisiontextrepresentationoptimizing}), +36.80 FB-IoU (SAN~\cite{xu2023adapternetworkopenvocabularysemantic}), and +2.47 hit@5 (MAFT+).

\paragraph{Zero-shot pose estimation.} We further evaluate \method’s ability to recover category-level 6D object poses from a single RGB image in a zero-shot setting. Due to space constraints, the full quantitative comparison with AG-Pose~\cite{lin2024instance}, SecondPose~\cite{Chen_2024_CVPR}, and MonoDiff9D~\cite{MonoDiff9D} across five indoor benchmarks is provided in the Supplementary. AG-Pose and SecondPose require RGB-D inputs, so we supply them with our predicted depths. Following MonoDiff9D, we report accuracy within 10 cm, within 10°, and under the joint 10°/10 cm criterion.

Across all datasets, \method shows consistent gains on the stricter angular and joint metrics. It attains the highest 10° and joint accuracy on PACE and HOPE and improves the joint metric on NOCS-real. For example, on PACE, \method improves the 10° rate from 15.1 to 25.9 and the joint 10°/10 cm rate from 8.6 to 14.0 compared to AG-Pose. These results indicate that the unified 3D representation learned by \method naturally supports precise, canonically aligned object poses without any dataset-specific finetuning.

In summary, our model's strong performance across monocular depth estimation, open-vocabulary semantic segmentation, and pose estimation jointly enables state-of-the-art 3D reconstruction quality, producing geometrically precise, semantically coherent, and canonically aligned scene representations.

\subsection{Ablation}
\label{sec:ablation}

\paragraph{Multi-task learning.} To demonstrate the advantage of using a unified model for multiple traditionally separated tasks, we retrain the model while removing one head at a time. Experiments shows that dropping semantic embeddings causes large pose and Gaussian degradations and also hurts depth. More broadly, removing any head weakens the remaining tasks, and the full four-head variant performs best. Full per-dataset quantitative results are provided in the Supplementary.

\begin{figure}[t]
    \centering
    \includegraphics[width=\columnwidth]{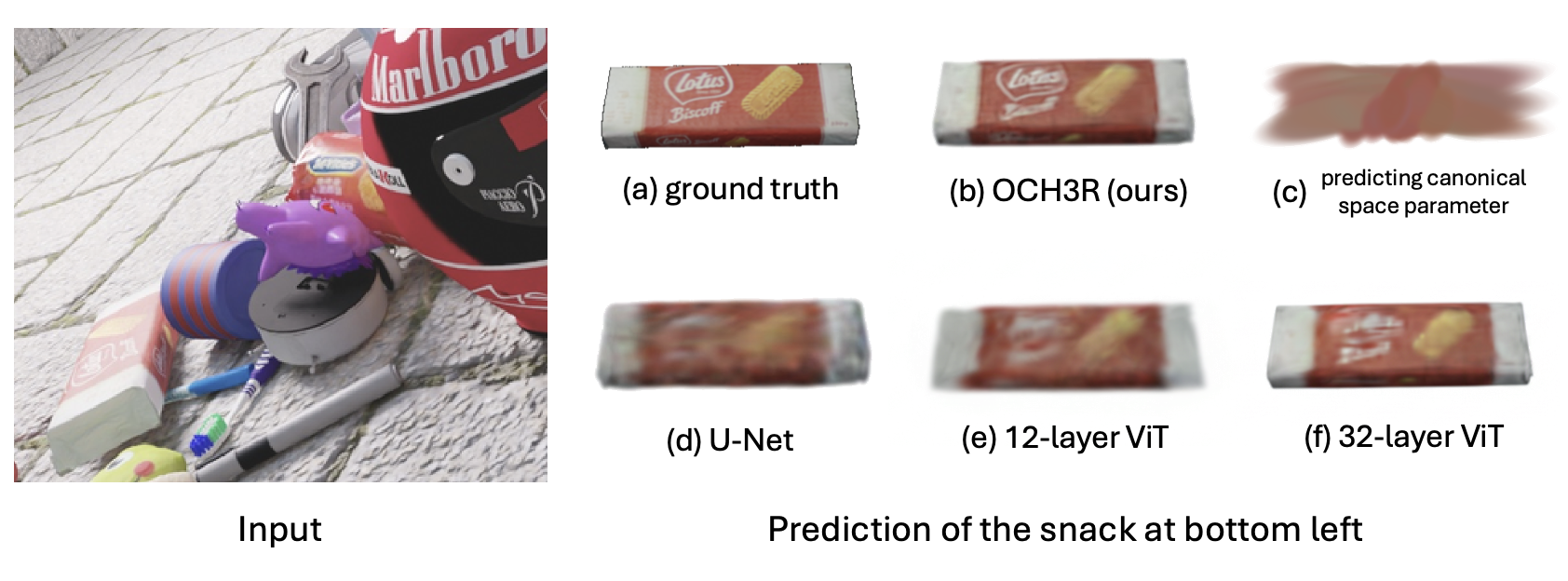}
    \caption{Qualitative ablations showing that (i) predicting Gaussians directly in canonical space collapses, and (ii) \method's formulation remains robust across model scales and architectures.}
    \label{fig:ablation}
    \vspace{-1.5em}
\end{figure}

\paragraph{Offset in camera space vs. directly in canonical space.} We also tried predicting Gaussian parameters directly in the canonical frame, bypassing the offset-along-ray formulation. As shown in \cref{fig:ablation}, removing the geometric scaffold of camera rays causes the network to collapse into an unstructured blob, confirming that unconstrained pixel-to-Gaussian mapping is highly underdetermined. By instead predicting per-pixel offsets in camera space and using Canonical-Space Supervision to organize the final layout, OCH3R obtains a stable and expressive inductive bias that enables high-quality reconstruction.

\paragraph{Model scale and architecture.} The core of our method, particularly the multi-task learning paradigm and Canonical-Space Supervision, is orthogonal to model architecture, so in principle any dense predictor could be used. We validate this by testing a U-Net (matched in parameter count to the 32-layer ViT) and two smaller ViT variants. As shown in \cref{fig:ablation}, while larger models yield sharper textures and cleaner geometry, all architectures successfully recover the object’s overall shape. This confirms that \method's pipeline transfers across dense predictors and scales well with model capacity.
\section{Conclusion}

In this paper, we try to address the long standing problem of understanding a scene as a set of discrete, posed objects from a single RGB image. Rather than relying on multi stage pipelines that decompose the task into segmentation, per object reconstruction, and post hoc alignment, we proposed \method, a unified, feed forward model that predicts all object instances, their category level $\mathrm{SIM}(3)$ poses, and high fidelity 3D Gaussians in a single pass. The key ingredients are a transformer that produces dense, pixel aligned attributes (metric depth, CLIP based semantics, NOCS coordinates, and per pixel Gaussians), a simple inference procedure for instance discovery and pose estimation, and canonical space supervision that trains amodally complete Gaussians without per image Gaussian labels.

To support this setting, we assembled a large scale dataset for holistic object centric 3D scene representation by aligning PACE, Omni6DPose, HOPE, YCB-Video and NOCS into a unified benchmark with per instance masks, semantics, 6D poses, and 3D models. Across this benchmark, \method outperforms previous baselines on monocular depth estimation, open vocabulary segmentation, and category level pose prediction, while producing more complete, editable 3D object reconstructions with feed forward inference that scales essentially independently of the number of objects.

\FloatBarrier

\section*{Acknowledgments}
This work used Marlowe~\cite{marlowe2025},
Stanford University's GPU-based Computational Instrument,
supported by Stanford HAI and Stanford Research Computing.

{
    \small
    \bibliographystyle{ieeenat_fullname}
    \bibliography{main}
}


\end{document}

\maketitlesupplementary

\begin{figure*}[ht]
    \centering
    \includegraphics[width=0.8\linewidth]{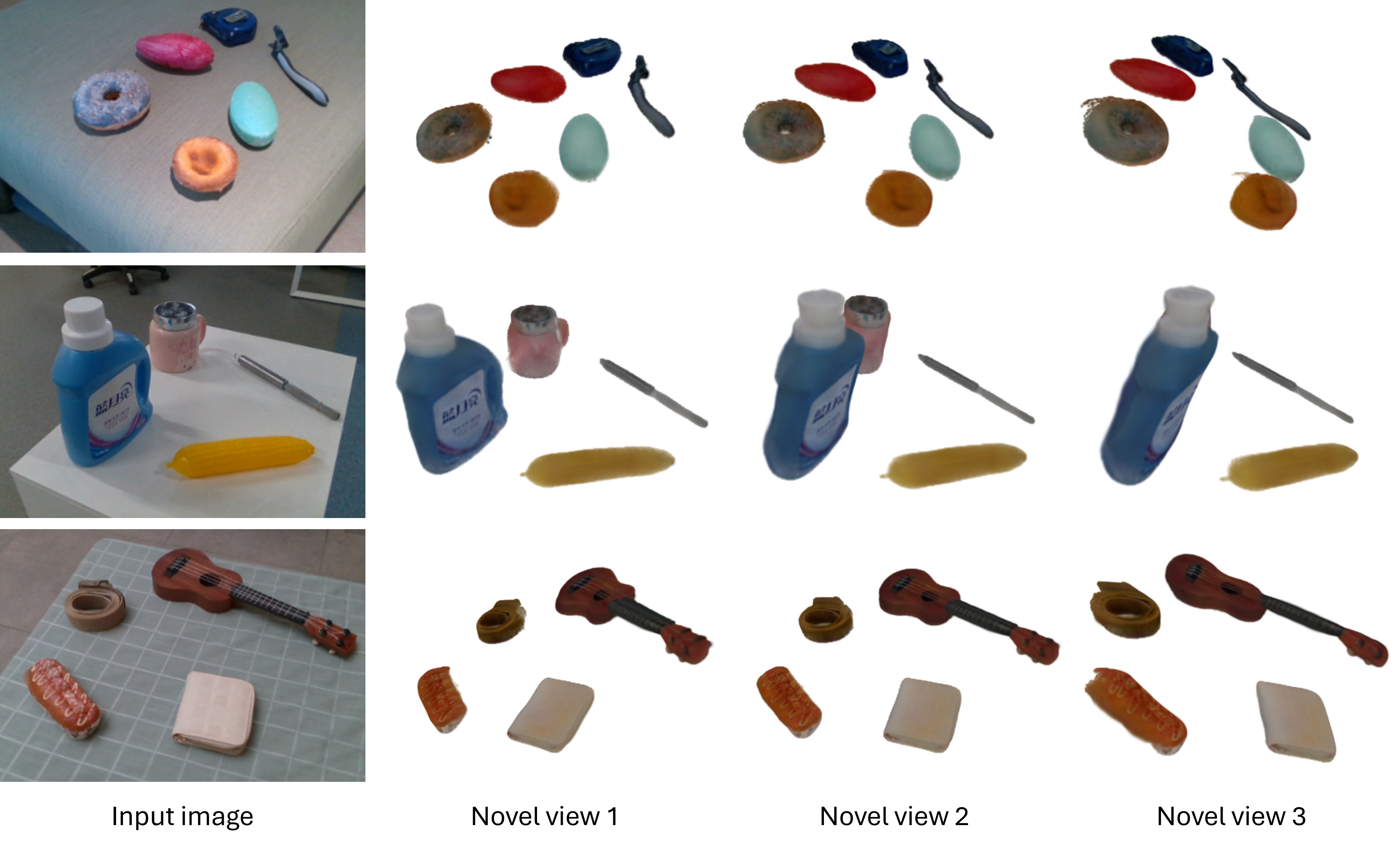}
    
    \caption{\textbf{Additional object-centric reconstruction results.} Each row corresponds to one input image (left, not shown here), and we render three novel views of the reconstructed scene from different camera poses. \method recovers coherent multi-object geometry that remains stable under large viewpoint changes.}
    \label{fig:more}
\end{figure*}

\begin{figure}
    \centering
    \includegraphics[width=\linewidth]{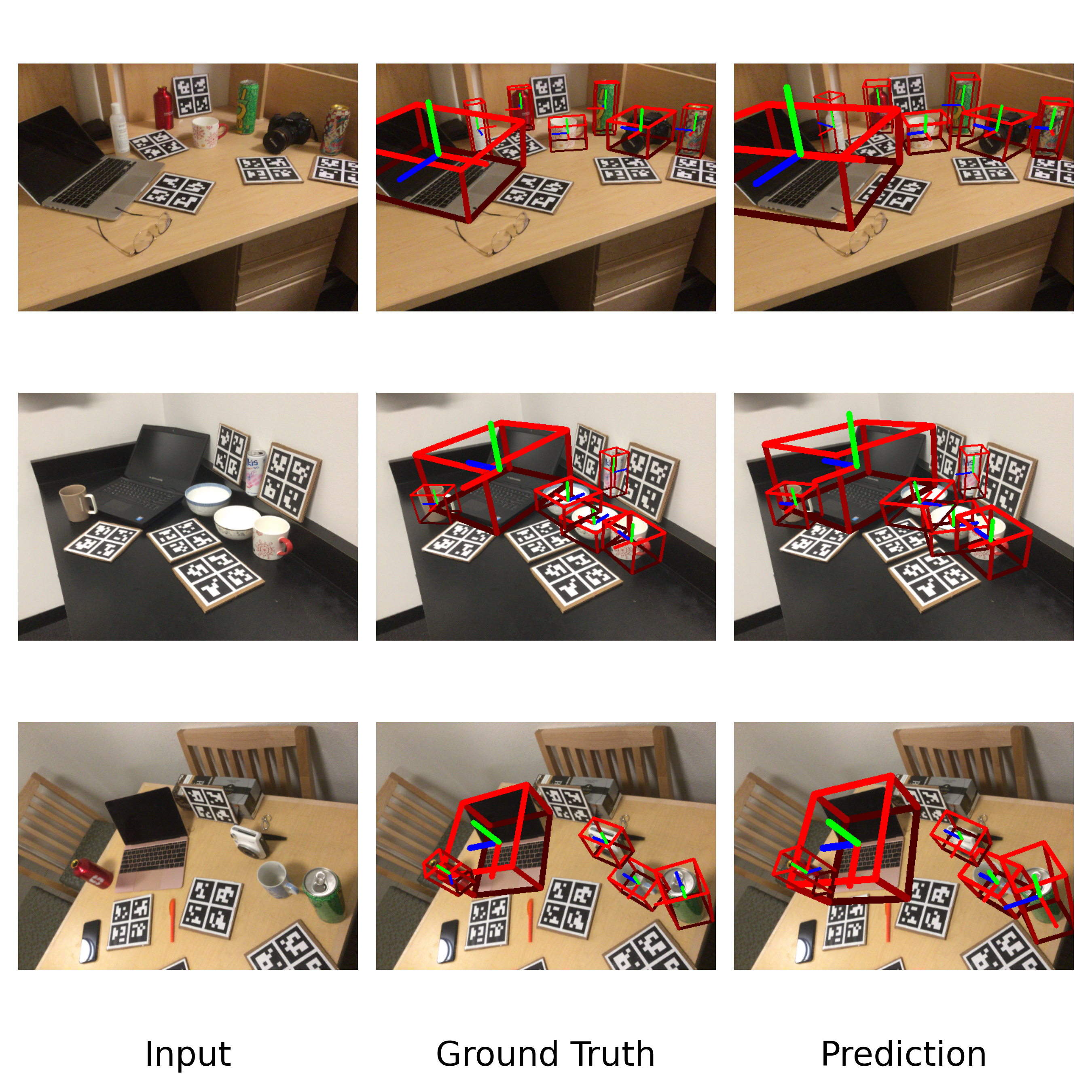}
    \caption{\textbf{Qualitative results for monocular pose estimation.} Our method is able to simultaneously predict the SIM(3) pose of all objects in the image.}
    \label{fig:pose_supp}
\end{figure}

\begin{table*}[h]
\centering
\small
\setlength{\tabcolsep}{4pt}
\resizebox{\textwidth}{!}{%
\begin{tabular}{lccccccccccccccc}
\toprule
\multirow{2}{*}{Model} 
& \multicolumn{3}{c}{PACE} 
& \multicolumn{3}{c}{OMNI} 
& \multicolumn{3}{c}{YCB-V} 
& \multicolumn{3}{c}{HOPE} 
& \multicolumn{3}{c}{NOCS~real} \\
\cmidrule(lr){2-4} \cmidrule(lr){5-7} \cmidrule(lr){8-10} \cmidrule(lr){11-13} \cmidrule(lr){14-16}
 & 10cm$\uparrow$ & $10^\circ$$\uparrow$ & $10^\circ$10cm$\uparrow$ 
 & 10cm$\uparrow$ & $10^\circ$$\uparrow$ & $10^\circ$10cm$\uparrow$
 & 10cm$\uparrow$ & $10^\circ$$\uparrow$ & $10^\circ$10cm$\uparrow$
 & 10cm$\uparrow$ & $10^\circ$$\uparrow$ & $10^\circ$10cm$\uparrow$
 & 10cm$\uparrow$ & $10^\circ$$\uparrow$ & $10^\circ$10cm$\uparrow$ \\
\midrule
AG-Pose~\cite{lin2024instance} 
& \textbf{44.2} & \underline{15.1} & \underline{8.6} 
& 0.0 & \underline{0.1} & 0.0 
& 0.0 & \textbf{16.2} & 0.0 
& \underline{38.8} & \underline{25.7} & \underline{9.7} 
& 48.7 & \underline{68.5} & 30.1 \\
SecondPose~\cite{Chen_2024_CVPR} 
& 40.7 & 14.6 & 8.4 
& 0.0 & \underline{0.1} & 0.0 
& 0.0 & 12.5 & 0.0 
& 38.6 & 23.6 & 8.4 
& \textbf{51.2} & 58.9 & \underline{31.6} \\
MonoDiff9D~\cite{MonoDiff9D} 
& 29.9 & 7.3 & 3.5 
& 0.0 & \underline{0.1} & 0.0 
& 0.0 & 7.9 & 0.0 
& 17.4 & 18.3 & 1.4 
& 41.0 & 56.3 & 25.7 \\
\midrule
\textbf{\method (ours)} 
& \underline{41.8} & \textbf{25.9} & \textbf{14.0} 
& \textbf{0.2} & \textbf{3.4} & 0.0 
& \textbf{0.7} & \underline{15.9} & 0.0 
& \textbf{48.3} & \textbf{28.5} & \textbf{11.1} 
& \underline{49.6} & \textbf{70.7} & \textbf{33.3} \\
\bottomrule
\end{tabular}
}
\caption{Zero-shot 6D pose estimation results. Metrics: percentage of predictions within 10cm, within $10^\circ$, and within both.}

\label{tab:pose_supp}
\end{table*}

\section{More reconstruction qualitative results}
\cref{fig:more} presents additional single-view reconstruction results, where we render each predicted object-centric scene from three held-out novel viewpoints per input image.
Across diverse indoor scenes, \method produces geometrically consistent reconstructions that maintain object shapes and relative spatial arrangements under large viewpoint changes.

\section{Additional individual task results}

% \begin{figure*}[h]
%     \centering
%     \includegraphics[width=0.9\linewidth]{figures/depth_qualitative.png}
%     \caption{\textbf{Qualitative results for zero-shot monocular metric depth estimation.} Our method produces crisp object boundaries, faithful fine-scale geometry, and globally accurate depth ordering, closely matching ground truth even under challenging occlusion, clutter, and lighting. Insets highlight sharp edges and well-defined local structures recovered.}
%     \label{fig:qual_tasks}
% \end{figure*}

\subsection{Pose estimation}
We compare our zero-shot pose estimation performance against state-of-the-art methods. As shown in \cref{tab:pose_supp}, \method achieves consistently strong results on standard accuracy metrics (accuracy within 10 cm, within 10°, and under the joint 10°/10 cm threshold) across multiple datasets, reflecting the effectiveness of our NOCS-based pose estimation formulation. The advantage stems from jointly reasoning about depth, semantics, and canonical correspondences, which provides strong geometric and semantic cues even under distribution shift.

In addition to the quantitative comparison, we also include qualitative visualization on held-out test data in \cref{fig:pose_supp}. These examples highlight that \method produces clean, well-aligned 6D poses even in cluttered or partially occluded scenarios.

\subsection{Full quantitative results for depth estimation}
Due to space constraints, the main paper reports only $\delta_1$, AbsRel, and RMSE for monocular depth evaluation. For completeness, \cref{tab:supp-depth-pace}--\cref{tab:supp-depth-nocs-real} provide the full set of standard depth metrics ($\delta_1$, $\delta_2$, $\delta_3$, AbsRel, $\log_{10}$, RMSE, RMSE$_{\log}$, SILog) for all methods and datasets. The additional metrics confirm the trends observed in the main paper: \method remains highly competitive, typically achieving the best or second-best performance across datasets and metrics.

\begin{table*}[t]
\centering
\small
\scalebox{0.85}{%
\begin{tabular}{lcccccccc}
\toprule
Model & $\delta_1 \, (\%) \, \uparrow$ & $\delta_2 \, (\%) \, \uparrow$ & $\delta_3 \, (\%) \, \uparrow$ & AbsRel $\downarrow$ & $\log_{10}$ $\downarrow$ & RMSE $\downarrow$ & RMSE$_{\log}$ $\downarrow$ & SILog $\downarrow$ \\
\midrule
NeWCRFs~\cite{yuan2022newcrfsneuralwindow} & 17.00 & 44.56 & 72.73 & 0.7356 & 0.2208 & 0.5308 & 0.5835 & 0.2952 \\
ZoeDepth~\cite{bhat2023zoedepthzeroshottransfercombining} & 8.20 & 31.99 & 62.11 & 0.8508 & 0.2530 & 0.6146 & 0.6366 & 0.2602 \\
Metric3D V2~\cite{Hu_2024} & 1.32 & 13.42 & 55.34 & 0.4695 & 0.2835 & 0.3995 & 0.6817 & \underline{0.1971} \\
Depth Anything V2~\cite{yang2024depthv2} & 13.07 & 37.82 & 70.65 & 0.7373 & 0.2274 & 0.5906 & 0.5789 & 0.2798 \\
Depth Pro~\cite{bochkovskii2025depthprosharpmonocular} & 33.63 & 61.60 & 81.87 & 0.5603 & 0.1728 & 0.5610 & 0.4989 & 0.3523 \\
VGGT~\cite{wang2025vggtvisualgeometrygrounded} & \underline{55.58} & \underline{91.24} & \underline{99.20} & \underline{0.2415} & \underline{0.0961} & \underline{0.2096} & \underline{0.2699} & 0.2607 \\
UniDepth V2~\cite{piccinelli2025unidepthv2universalmonocularmetric} & 35.05 & 63.91 & 83.87 & 0.5076 & 0.1667 & 0.4820 & 0.4783 & 0.3896 \\
\midrule
\textbf{\method (ours)} & \textbf{94.82} & \textbf{99.60} & \textbf{99.88} & \textbf{0.0834} & \textbf{0.0368} & \textbf{0.1039} & \textbf{0.1168} & \textbf{0.1142} \\
\bottomrule
\end{tabular}
}
\vspace{-0.7em}
\caption{Full monocular depth estimation results on PACE (test split). $\delta_1$, $\delta_2$, and $\delta_3$ are reported in percentage. All models are evaluated using their largest available variant.}
\label{tab:supp-depth-pace}
\vspace{6pt}
\centering
\small
\scalebox{0.85}{%
\begin{tabular}{lcccccccc}
\toprule
Model & $\delta_1 \, (\%) \, \uparrow$ & $\delta_2 \, (\%) \, \uparrow$ & $\delta_3 \, (\%) \, \uparrow$ & AbsRel $\downarrow$ & $\log_{10}$ $\downarrow$ & RMSE $\downarrow$ & RMSE$_{\log}$ $\downarrow$ & SILog $\downarrow$ \\
\midrule
NeWCRFs~\cite{yuan2022newcrfsneuralwindow} & 21.41 & 50.42 & 78.75 & 0.6432 & 0.2015 & 0.9934 & 0.5416 & 0.3429 \\
ZoeDepth~\cite{bhat2023zoedepthzeroshottransfercombining} & 20.73 & 54.47 & 84.47 & 0.5794 & 0.1881 & 0.9406 & 0.4961 & 0.2825 \\
Metric3D V2~\cite{Hu_2024} & \textbf{73.73} & \textbf{94.56} & \underline{98.85} & \textbf{0.1788} & \textbf{0.0740} & \underline{0.7963} & \textbf{0.2371} & \textbf{0.2349} \\
Depth Anything V2~\cite{yang2024depthv2} & 5.83 & 27.10 & 61.84 & 0.9009 & 0.2651 & 1.1728 & 0.6641 & 0.2823 \\
Depth Pro~\cite{bochkovskii2025depthprosharpmonocular} & \underline{57.37} & 86.69 & 96.35 & 0.2742 & 0.1023 & \textbf{0.7940} & 0.3108 & \underline{0.2789} \\
VGGT~\cite{wang2025vggtvisualgeometrygrounded} & 57.14 & 90.28 & 97.98 & \underline{0.2272} & 0.0992 & 1.0365 & 0.3050 & 0.3017 \\
UniDepth V2~\cite{piccinelli2025unidepthv2universalmonocularmetric} & 22.80 & 50.17 & 73.47 & 0.7221 & 0.2155 & 1.1007 & 0.5997 & 0.4303 \\
\midrule
\textbf{\method (ours)} & 42.77 & \underline{91.76} & \textbf{99.23} & 0.2900 & \underline{0.0812} & 0.8042 & \underline{0.2743} & 0.2897 \\
\bottomrule
\end{tabular}
}
\vspace{-0.7em}
\caption{Full monocular depth estimation results on OMNI (Rope split). $\delta_1$, $\delta_2$, and $\delta_3$ are reported in percentage. All models are evaluated using their largest available variant.}
\label{tab:supp-depth-omni-rope}
\vspace{6pt}
\centering
\small
\scalebox{0.85}{%
\begin{tabular}{lcccccccc}
\toprule
Model & $\delta_1 \, (\%) \, \uparrow$ & $\delta_2 \, (\%) \, \uparrow$ & $\delta_3 \, (\%) \, \uparrow$ & AbsRel $\downarrow$ & $\log_{10}$ $\downarrow$ & RMSE $\downarrow$ & RMSE$_{\log}$ $\downarrow$ & SILog $\downarrow$ \\
\midrule
NeWCRFs~\cite{yuan2022newcrfsneuralwindow} & 52.71 & 91.70 & 99.13 & 0.2740 & 0.1004 & 0.2699 & 0.2759 & 0.1687 \\
ZoeDepth~\cite{bhat2023zoedepthzeroshottransfercombining} & 52.84 & 90.67 & 98.85 & 0.2704 & 0.0986 & 0.2750 & 0.2764 & 0.1727 \\
Metric3D V2~\cite{Hu_2024} & 15.44 & 84.96 & \underline{99.82} & 0.2791 & 0.1443 & 0.2538 & 0.3486 & \textbf{0.1107} \\
Depth Anything V2~\cite{yang2024depthv2} & 50.74 & \textbf{97.23} & 99.62 & \underline{0.1899} & 0.0929 & \textbf{0.2029} & \underline{0.2532} & 0.1709 \\
Depth Pro~\cite{bochkovskii2025depthprosharpmonocular} & 49.04 & 94.40 & 99.65 & 0.2078 & 0.1000 & 0.2148 & 0.2714 & 0.2212 \\
VGGT~\cite{wang2025vggtvisualgeometrygrounded} & \underline{63.88} & 87.46 & 97.93 & \textbf{0.1822} & \underline{0.0900} & 0.2131 & 0.2742 & 0.2468 \\
UniDepth V2~\cite{piccinelli2025unidepthv2universalmonocularmetric} & 32.70 & 80.73 & 99.43 & 0.2548 & 0.1315 & 0.2499 & 0.3416 & 0.1783 \\
\midrule
\textbf{\method (ours)} & \textbf{69.96} & \underline{95.34} & \textbf{99.87} & 0.1933 & \textbf{0.0864} & \underline{0.2044} & \textbf{0.2405} & \underline{0.1635} \\
\bottomrule
\end{tabular}
}
\vspace{-0.7em}
\caption{Full monocular depth estimation results on YCB-V. $\delta_1$, $\delta_2$, and $\delta_3$ are reported in percentage. All models are evaluated using their largest available variant.}
\label{tab:supp-depth-ycbv}
\vspace{6pt}
\centering
\small
\scalebox{0.85}{%
\begin{tabular}{lcccccccc}
\toprule
Model & $\delta_1 \, (\%) \, \uparrow$ & $\delta_2 \, (\%) \, \uparrow$ & $\delta_3 \, (\%) \, \uparrow$ & AbsRel $\downarrow$ & $\log_{10}$ $\downarrow$ & RMSE $\downarrow$ & RMSE$_{\log}$ $\downarrow$ & SILog $\downarrow$ \\
\midrule
NeWCRFs~\cite{yuan2022newcrfsneuralwindow} & 5.37 & 33.98 & 64.47 & 0.8073 & 0.2458 & 0.5029 & 0.6107 & 0.2335 \\
ZoeDepth~\cite{bhat2023zoedepthzeroshottransfercombining} & 0.53 & 21.32 & 63.87 & 0.9814 & 0.2812 & 0.6055 & 0.6990 & 0.2667 \\
Metric3D V2~\cite{Hu_2024} & 0.01 & 24.72 & 91.66 & 0.4039 & 0.2274 & 0.2968 & 0.5364 & \textbf{0.1167} \\
Depth Anything V2~\cite{yang2024depthv2} & 0.07 & 0.24 & 49.05 & 1.0364 & 0.3055 & 0.6518 & 0.7147 & 0.1340 \\
Depth Pro~\cite{bochkovskii2025depthprosharpmonocular} & 23.09 & \underline{82.05} & \underline{99.79} & \underline{0.4035} & \underline{0.1437} & 0.3013 & \underline{0.3581} & 0.1437 \\
VGGT~\cite{wang2025vggtvisualgeometrygrounded} & \underline{37.17} & 72.29 & 84.27 & 0.4313 & 0.1438 & \underline{0.2844} & 0.4043 & 0.2439 \\
UniDepth V2~\cite{piccinelli2025unidepthv2universalmonocularmetric} & 8.63 & 47.83 & 84.67 & 0.7622 & 0.2190 & 0.6830 & 0.5931 & 0.3173 \\
\midrule
\textbf{\method (ours)} & \textbf{61.36} & \textbf{98.82} & \textbf{99.91} & \textbf{0.2192} & \textbf{0.0852} & \textbf{0.1812} & \textbf{0.2351} & \underline{0.1296} \\
\bottomrule
\end{tabular}
}
\vspace{-0.7em}
\caption{Full monocular depth estimation results on HOPE. $\delta_1$, $\delta_2$, and $\delta_3$ are reported in percentage. All models are evaluated using their largest available variant.}
\label{tab:supp-depth-hope}
\vspace{6pt}
\centering
\small
\scalebox{0.85}{%
\begin{tabular}{lcccccccc}
\toprule
Model & $\delta_1 \, (\%) \, \uparrow$ & $\delta_2 \, (\%) \, \uparrow$ & $\delta_3 \, (\%) \, \uparrow$ & AbsRel $\downarrow$ & $\log_{10}$ $\downarrow$ & RMSE $\downarrow$ & RMSE$_{\log}$ $\downarrow$ & SILog $\downarrow$ \\
\midrule
NeWCRFs~\cite{yuan2022newcrfsneuralwindow} & 36.00 & 86.70 & 99.40 & 0.3352 & 0.1211 & 0.3586 & 0.3133 & 0.1503 \\
ZoeDepth~\cite{bhat2023zoedepthzeroshottransfercombining} & 34.08 & 66.71 & 93.76 & 0.4399 & 0.1492 & 0.5275 & 0.3993 & 0.2090 \\
Metric3D V2~\cite{Hu_2024} & 8.50 & 60.26 & 99.66 & 0.3273 & 0.1747 & 0.3571 & 0.4160 & \underline{0.1068} \\
Depth Anything V2~\cite{yang2024depthv2} & 2.44 & 69.03 & 90.58 & 0.5507 & 0.1851 & 0.6088 & 0.4533 & 0.1554 \\
Depth Pro~\cite{bochkovskii2025depthprosharpmonocular} & 81.58 & \underline{99.91} & \underline{99.99} & 0.1370 & 0.0545 & 0.1831 & 0.1576 & 0.1322 \\
VGGT~\cite{wang2025vggtvisualgeometrygrounded} & \underline{92.26} & 99.66 & 99.98 & \underline{0.0965} & \underline{0.0438} & \underline{0.1394} & \underline{0.1278} & 0.1200 \\
UniDepth V2~\cite{piccinelli2025unidepthv2universalmonocularmetric} & 78.92 & 97.17 & 99.16 & 0.1778 & 0.0678 & 0.2366 & 0.1979 & 0.1339 \\
\midrule
\textbf{\method (ours)} & \textbf{98.43} & \textbf{99.97} & \textbf{100.00} & \textbf{0.0923} & \textbf{0.0378} & \textbf{0.1066} & \textbf{0.1051} & \textbf{0.0704} \\
\bottomrule
\end{tabular}
}
\vspace{-0.7em}
\caption{Full monocular depth estimation results on NOCS (real split). $\delta_1$, $\delta_2$, and $\delta_3$ are reported in percentage. All models are evaluated using their largest available variant.}
\label{tab:supp-depth-nocs-real}
\end{table*}

\section{Network and training details}
\subsection{Network architecture}
Our architecture is inspired by VGGT~\cite{wang2025vggtvisualgeometrygrounded} and builds on the DINOv2 ViT-L/14~\cite{oquab2024dinov2learningrobustvisual} patch embedding as the backbone initialization. The input image is split into non-overlapping $14\times14$ patches, and each patch is linearly projected into a token. The resulting token sequence is then processed by a 42-layer transformer encoder with an embedding dimension of 1024 and 16 attention heads.

We employ four lightweight DPT-style~\cite{ranftl2021visiontransformersdenseprediction} decoder heads for dense per-pixel prediction, together with a separate global camera head:
\begin{itemize}
    \item \textbf{Semantic Head}: Consumes multi-scale encoder features and fuses them into a full-resolution feature map using lateral skips and convolutional fusion. A small projection MLP (hidden dimension 2048) maps these features to 512-dimensional per-pixel embeddings that live in the same space as CLIP text embeddings. During training, we compute cosine similarities between pixel embeddings and a set of CLIP text embeddings and use them as logits for an open-vocabulary semantic segmentation loss.

    \item \textbf{Depth Head}: Shares the same DPT-style fusion as the semantic head but outputs a single-channel canonical inverse-depth map $\hat{C}_{u,v}$. Following DepthPro-style canonicalization, we supervise $\hat{C}$ via $C = f_w / (W \cdot d)$, where $f_w$ is the horizontal focal length in pixels, $W$ is the image width, and $d$ is metric depth. At inference time, we recover metric depth by combining $\hat{C}$ with the focal length derived from the predicted field-of-view.

    \item \textbf{NOCS Head}: Predicts per-pixel NOCS maps, i.e., 3D coordinates in a category-level canonical unit cube with consistent orientation and scale~\cite{wang2019normalized}. To better handle the multi-modality caused by object symmetries, we adopt a mixed classification–regression ``bin-and-delta'' parameterization: each of the $X$, $Y$, and $Z$ coordinates is discretized into 64 uniform bins over $[0,1]$, and the network jointly predicts a 64-way categorical distribution over bins together with a continuous offset from the chosen bin center. Prior work in NOCS-based reconstruction and category-level pose estimation has shown that discretizing continuous coordinates into bins provides greater robustness than direct $L_2$ regression under symmetric ambiguities~\cite{wang2019normalized,chi2021garmentnetscategorylevelposeestimation,xue2025garmenttrackingcategorylevelgarmentpose}. Our formulation aligns with this strategy and follows the broader “bin-and-delta’’ design widely adopted in 3D pose estimation~\cite{li2018unifiedframeworkmultiviewmulticlass,levinson2020analysissvddeeprotation}.

    \item \textbf{Gaussian Head}: Implements our feed-forward 3D Gaussian predictor and is split into two sub-heads that share the same DPT-style multi-scale features but use a learnable upsampler (PixelShuffle or ConvTranspose-based) instead of simple bilinear interpolation to avoid over-smoothing structured parameters. The \emph{geometry} sub-head predicts, for each pixel and for each of $k{=}2$ Gaussians, an off-ray offset $\boldsymbol{\Delta}^{(i)}_{u,v} \in \mathbb{R}^3$ in camera space; adding this to the back-projected 3D point $\hat{\mathbf p}_{u,v}$ yields the Gaussian means $\boldsymbol{\mu}^{(i)}_{u,v}$ in the camera frame. The \emph{appearance/shape} sub-head predicts the remaining parameters in the canonical frame: log-scales (3 dims), a unit quaternion (4 dims) that defines the Gaussian orientation, opacity $\alpha^{(i)}_{u,v}$ (scalar), and RGB Spherical Harmonics coefficients of degree 0 (3 dims per Gaussian). We concatenate all parameters for the $k$ Gaussians at each pixel into a single output tensor. Following NoPoSplat-style designs, this head also receives a shallow RGB shortcut to better preserve fine texture details in the predicted Gaussian colors.

    \item \textbf{Camera Head}: In addition to the dense heads, we attach a global camera head that follows the VGGT formulation. A learnable camera token is appended to the patch tokens and processed by a small stack of transformer blocks with adaptive layer-norm modulation and iterative refinement. A linear projection from the final camera token regresses the horizontal and vertical field-of-view angles $(\hat{\theta}_w, \hat{\theta}_h)$ of the input image. These angles are converted into focal lengths via $f_w = \tfrac{W}{2 \tan(\hat{\theta}_w / 2)}$ and $f_h = \tfrac{H}{2 \tan(\hat{\theta}_h / 2)}$, which define the intrinsics matrix $K$ used by our depth, NOCS, and Gaussian heads.
\end{itemize}

\subsection{Training time and learning rate}
We implement our model in PyTorch using PyTorch Lightning. Our model contains 1.4B parameters. Training is performed on 32 NVIDIA H100 GPUs (4 nodes $\times$ 8 GPUs) for 30 epochs. The effective batch size is 64 (2 per GPU). We use the AdamW optimizer with a weight decay of 0.05. The learning rate is set to $1.25 \times 10^{-4}$ with a linear warmup for the first 5\% of steps, followed by a cosine decay schedule to a minimum of $1 \times 10^{-6}$.

\subsection{Data augmentation}
During training, we resize all images to a fixed width of $518$. Following VGGT, we randomly crop the image with different aspect ratios ranging from 0.33 to 1. We apply data augmentation to enhance diversity, specifically utilizing random color jittering with a probability of 0.9 (adjusting brightness, contrast, and saturation by a factor of 0.5, and hue by 0.1), followed by random grayscale conversion with a probability of 0.05. We do not use horizontal flipping to maintain consistency in the NOCS coordinate system.

\subsection{Homoscedastic uncertainty weighting}
We jointly optimize all tasks (depth, semantics, NOCS, Gaussian reconstruction, and camera FOV) using homoscedastic uncertainty weighting~\cite{kendall2018multitasklearningusinguncertainty}. Concretely, let $\mathcal{L}_{\text{depth}}, \mathcal{L}_{\text{sem}}, \mathcal{L}_{\text{nocs}}, \mathcal{L}_{\text{css}}, \mathcal{L}_{\text{cam}}$ denote the per-task losses defined in Sec.~4 of the main paper. We associate each task $t$ with a learnable log-variance parameter $s_t \in \mathbb{R}$, which is shared across all training examples and does not depend on the input (i.e., it models homoscedastic, task-specific observation noise). The total loss is
\begin{equation}
\mathcal{L}_{\text{total}}
= \sum_{t \in \{\text{depth}, \text{sem}, \text{nocs}, \text{css}, \text{cam}\}}
\bigl( \exp(-s_t)\, \mathcal{L}_t + s_t \bigr),    
\end{equation}
which is equivalent (up to constants) to the maximum-likelihood formulation in~\cite{kendall2018multitasklearningusinguncertainty}. This parameterization guarantees positive effective weights $\exp(-s_t)$ and allows the model to automatically balance the relative importance of each task, avoiding manual tuning of heuristic loss coefficients while remaining stable throughout training.

% \subsection{CRF configuration}
% At inference time, we refine the raw per-pixel CLIP embeddings with a fully connected CRF to obtain coherent instance masks. We employ a DenseCRF~\cite{krahenbuhl2011efficient} formulation where unary terms are computed as the negative log-softmax over the category logits derived from CLIP similarities (temperature $\tau = 0.07$). The pairwise potential is composed of two Gaussian kernels: a feature-dependent bilateral kernel driven by the embedding cosine similarity (weight 10.0, spatial $\sigma =20$, feature $\sigma =0.1$) to capture the semantic affinities, and a spatial smoothness kernel (weight 3.0, spatial $\sigma=3$) to enforce local continuity. We run 5 mean-field iterations using a CUDA implementation, which adds roughly 200 ms per image at our resolution of 518 pixels in width. The final CRF labels are used to define object masks for downstream pose estimation and Gaussian aggregation and are not used during training.

\section{Dataset details}
\subsection{Labeling and alignment}
For semantic segmentation, we curated a unified vocabulary of 234 everyday object categories tailored specifically for our training setup. Rather than directly adopting the category sets from PACE~\cite{you2024pacelargescaledatasetpose}, Omni6DPose~\cite{zhang2024omni6dposebenchmarkmodeluniversal}, GSO~\cite{downs2022googlescannedobjectshighquality}, and HyperSim~\cite{roberts2021hypersimphotorealisticsyntheticdataset}, we consolidated and cleaned their ontologies, removed redundancies, harmonized naming conventions, and added missing but frequently observed items. This ensured comprehensive coverage of objects across all four datasets while maintaining a clean, consistent taxonomy suitable for multi-task training. The full vocabulary used for training is listed in \cref{tab:categories}. We use the pre-trained CLIP text encoder to obtain a 512-dimensional embedding for each category name.

For 3D alignment, we standardize object geometry and pose across heterogeneous sources to enable consistent supervision. PACE and Omni6DPose provide canonical 6D object poses out of the box. For GSO and HyperSim, however, we perform an additional normalization and alignment step: we center each asset, scale it to a unit cube following the NOCS convention, and manually correct orientation inconsistencies that arise due to varying dataset conventions and asset preparation pipelines. This careful consolidation produces a coherent canonical frame across all datasets, allowing us to supervise the NOCS head and the canonical-space Gaussian predictions under a uniform geometric definition.

\subsection{Dataset Statistics}
\begin{figure*}[htb]
    \centering
    \includegraphics[width=0.9\linewidth]{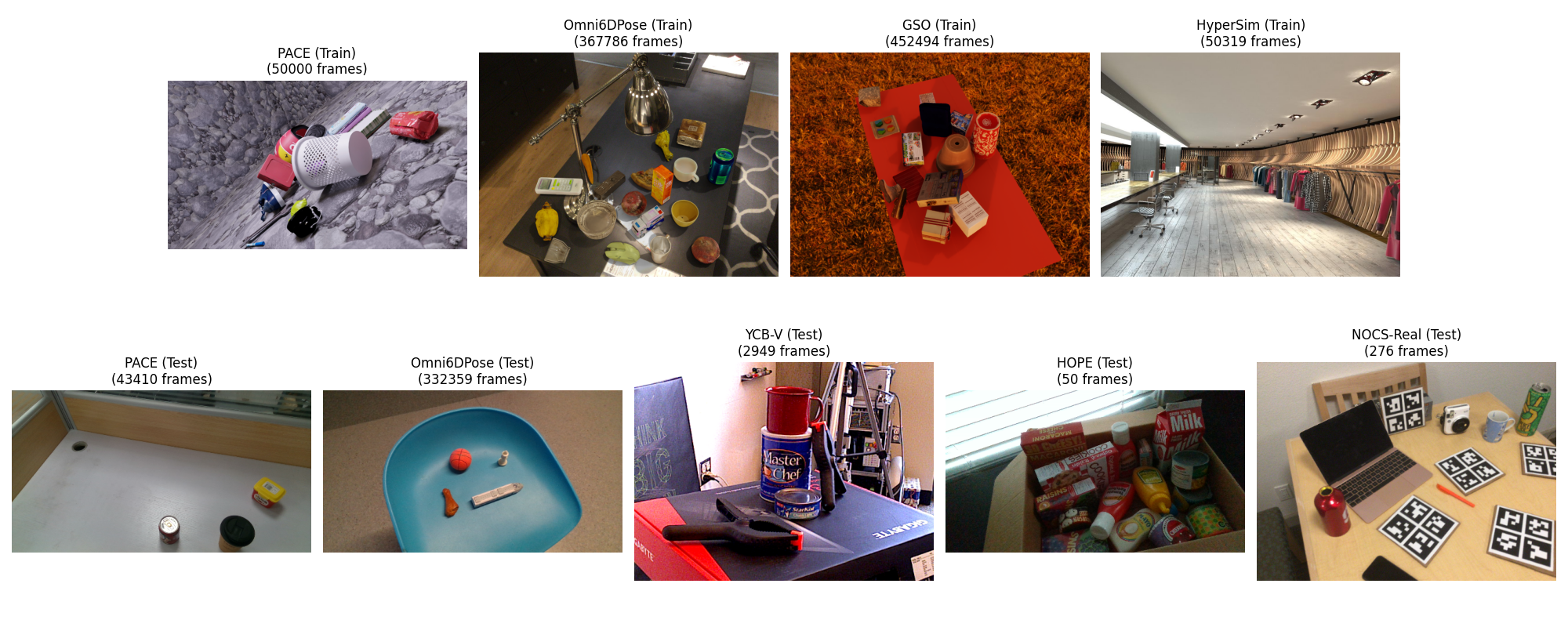}
    
    \caption{\textbf{Training and test datasets.} Representative RGB images from the four training datasets (PACE, Omni6DPose, GSO, HyperSim; top row) and five test datasets (PACE, Omni6DPose, YCB-V, HOPE, NOCS-Real; bottom row).}
    \label{fig:dataset_samples}
\end{figure*}

Our training corpus combines 50{,}000 frames from the PACE-Sim training split, 367{,}786 frames from the Omni6DPose SOPE\_train split, 452{,}494 rendered views from Google Scanned Objects (GSO), and 50{,}319 frames from the HyperSim training split, for a total of 920{,}599 training images.
For evaluation, we use 43{,}410 frames from the PACE test split, 332{,}359 frames from the Omni6DPose ROPE\_test split, 2{,}949 keyframes from the YCB-V test\_keyframe split, 50 validation images from HOPE, and 276 images from the NOCS-Real test split.
\cref{fig:dataset_samples} shows representative examples from each training and test dataset.

\section{Ablation}
\subsection{Multi-task learning ablation on PACE}

In Sec.~6.3 of the main paper we ablate the unified four-head design by removing one prediction head at a time (semantics, depth, NOCS, Gaussians).
\cref{tab:ablation_multi_task} shows full results on the PACE validation split. We report depth accuracy $\delta_1$ (higher is better), pose accuracy under the joint $10^\circ$/10\,cm criterion A$_{10^\circ/10\text{cm}}$, and Gaussian reconstruction PSNR (dB) in canonical space.

The full model with all four heads attains the best performance across all three metrics, indicating that the joint training objective provides a strong inductive bias for both geometry and semantics. Removing the semantic head leads to the largest drop in pose and PSNR, suggesting that semantic embeddings are particularly important for stabilizing pose estimation and Gaussian reconstruction. Dropping the depth head also degrades pose and PSNR, showing that metric depth supervision benefits downstream 6D pose and canonical-space reconstruction even when depth is not evaluated directly. Omitting the NOCS head harms both depth and PSNR, highlighting the role of canonical coordinates in regularizing the 3D layout. Finally, training without the Gaussian head slightly reduces depth and pose accuracy, indicating that the reconstruction loss feeds back into the shared backbone and improves the other tasks as well.

\begin{table}[H]
\centering
\small
\setlength{\tabcolsep}{6pt}
\begin{tabular}{lccc}
\toprule
Model & $\delta_1 \uparrow$ & A$_{10^\circ/10\text{cm}} \uparrow$ & PSNR (dB)$\uparrow$ \\
\midrule
Full (all heads)      & \textbf{95.6} & \textbf{14.2} & \textbf{28.6} \\
w/o Sem. head         & 91.1 &  8.9 & 24.0 \\
w/o Depth head        &  --  & 11.6 & 27.8 \\
w/o NOCS head         & 88.2 &  --  & 25.2 \\
w/o Gauss head        & 95.2 & 13.5 &  --  \\
\bottomrule
\end{tabular}
\caption{\textbf{Multi-task ablation} on PACE (validation split). Depth ($\delta_1$, \%), pose accuracy A$_{10^\circ/10\text{cm}}$ (\%), and Gaussian PSNR (dB) when removing one prediction head at a time. For PSNR, we transform the predicted Gaussians into canonical space and render them using ground-truth instance information for all non-Gaussian quantities, so the metric isolates the effect of the Gaussian head. }
\label{tab:ablation_multi_task}
\end{table}

\subsection{Offset in camera space vs. direct canonical Gaussians}

In Sec.~6.3 of the main paper we qualitatively compared our default
camera-space offset parameterization of Gaussian means with a variant
that predicts Gaussian parameters directly in the canonical frame, bypassing
the offset-along-ray formulation. For completeness, we report
quantitative results of this ablation.

The {\bf camera-space} model is our default \method: each pixel first predicts
metric depth and a per-pixel 3D point along the camera ray, and the Gaussian
means are obtained by adding a learned off-ray offset in camera space before
being transformed into canonical space via the object $\mathrm{SIM}(3)$ pose. The {\bf direct canonical} ablation instead predicts Gaussian means directly in the canonical frame from the same backbone features; these Gaussians are rendered in canonical space with CSS, but no geometric constraint ties a given pixel to a particular region of the object.

We train both variants from scratch with identical hyperparameters. For geometry, we follow the evaluation protocol of Sec.~6.1 and report
Chamfer Distance and F-1@0.1 between the reconstructed and ground-truth meshes. For appearance, we measure CLIP similarity between canonical renderings of the predicted and ground-truth objects, and additionally report PSNR (in dB) computed over the same set of canonical views used for Canonical-Space Supervision.

\begin{table}[H]
\centering
\small
\resizebox{\linewidth}{!}{
\begin{tabular}{lcccc}
\toprule
Model & CD$\downarrow$ & F-1@0.1$\uparrow$ & CLIP$\uparrow$ & PSNR$\uparrow$ (dB) \\
\midrule
Camera-space offsets (ours)   & \textbf{0.18} & \textbf{45.90} & \textbf{83.41} & \textbf{27.57} \\
Direct canonical Gaussians    & 0.35 & 15.10 & 79.20 & 23.65 \\
\bottomrule
\end{tabular}
}
\caption{\textbf{Camera-space offsets vs.\ direct canonical Gaussians} on
PACE (validation split). Predicting Gaussian means as camera-space
offsets along rays yields substantially better geometry and appearance
than regressing Gaussians directly in canonical space.}
\label{tab:pace_offset_vs_canonical}
\end{table}

On PACE (validation split), removing the camera-ray scaffold causes the network to
converge to a degenerate solution: average Chamfer Distance increases
from 0.18 to 0.35, and F-1@0.1 falls from 45.90 to
15.10. Despite this geometric collapse, the drop in CLIP
similarity is comparatively modest (from 83.4 to 79.2),
indicating that the direct-canonical model still matches coarse color
and texture but fails to organize Gaussians into a coherent 3D shape.
The strong PSNR gap in canonical space (27.57 dB vs. 23.65 dB)
confirms that camera-space offsets provide a crucial inductive bias:
they tie each pixel’s Gaussians to a geometrically meaningful 3D
location, allowing CSS to assemble accurate and
amodally complete reconstructions.

\subsection{Effect of model scale and architecture}
\label{sec:supp:model-scale}

As discussed in the main paper, our formulation, particularly the multi-task learning setup and Canonical-Space Supervision, is largely orthogonal to the choice of dense backbone. To quantify this, we evaluate three transformer backbones with different depths (12, 32, and 48 layers) as well as a U-Net whose parameter count roughly matches the 32-layer ViT. All models share the same DINOv2~\cite{oquab2024dinov2learningrobustvisual} image encoder and task heads and are trained on the same data. \cref{tab:ablation_model_scale} reports results on the PACE validation split, using the 3D reconstruction metrics from the main paper: Chamfer Distance (CD), F-1@0.1, and CLIP similarity between rendered predictions and ground-truth images, along with average inference time per image.

\begin{table}[H]
  \centering
  \setlength{\tabcolsep}{7pt}
  \small
  \begin{tabular}{lcccc}
    \toprule
    Model & CD$\downarrow$ & F-1$\uparrow$ & CLIP$\uparrow$ & Time / img (s)$\downarrow$ \\
    \midrule
    ViT-12           & 0.30 & 28.93 & 81.05 & \textbf{0.45} \\
    U-Net      & 0.29 & 31.11 & 81.88 & 0.52 \\
    ViT-32           & 0.20 & 40.05 & 82.36 & 0.59 \\
    ViT-48 (ours)  & \textbf{0.18} & \textbf{45.90} & \textbf{83.41} & 0.70 \\
    \bottomrule
  \end{tabular}
  \caption{\textbf{Model scale and architecture ablation} on PACE (validation split). CD is Chamfer Distance (lower is better), F-1 is evaluated at $0.1$,
  and CLIP is image-space similarity (higher is better).}
  \label{tab:ablation_model_scale}
\end{table}

Across all metrics, performance scales smoothly with model capacity:
the smallest ViT-12 backbone already yields strong reconstructions, while
the 48-layer ViT further reduces CD and improves F-1 and CLIP, at the
cost of slightly higher runtime. The U-Net remains competitive with the
transformer models, supporting our claim that the \method pipeline
transfers across dense architectures while benefiting from additional
capacity.

\section{Failure cases and limitations}
\begin{figure}
    \centering
    \includegraphics[width=0.9\linewidth]{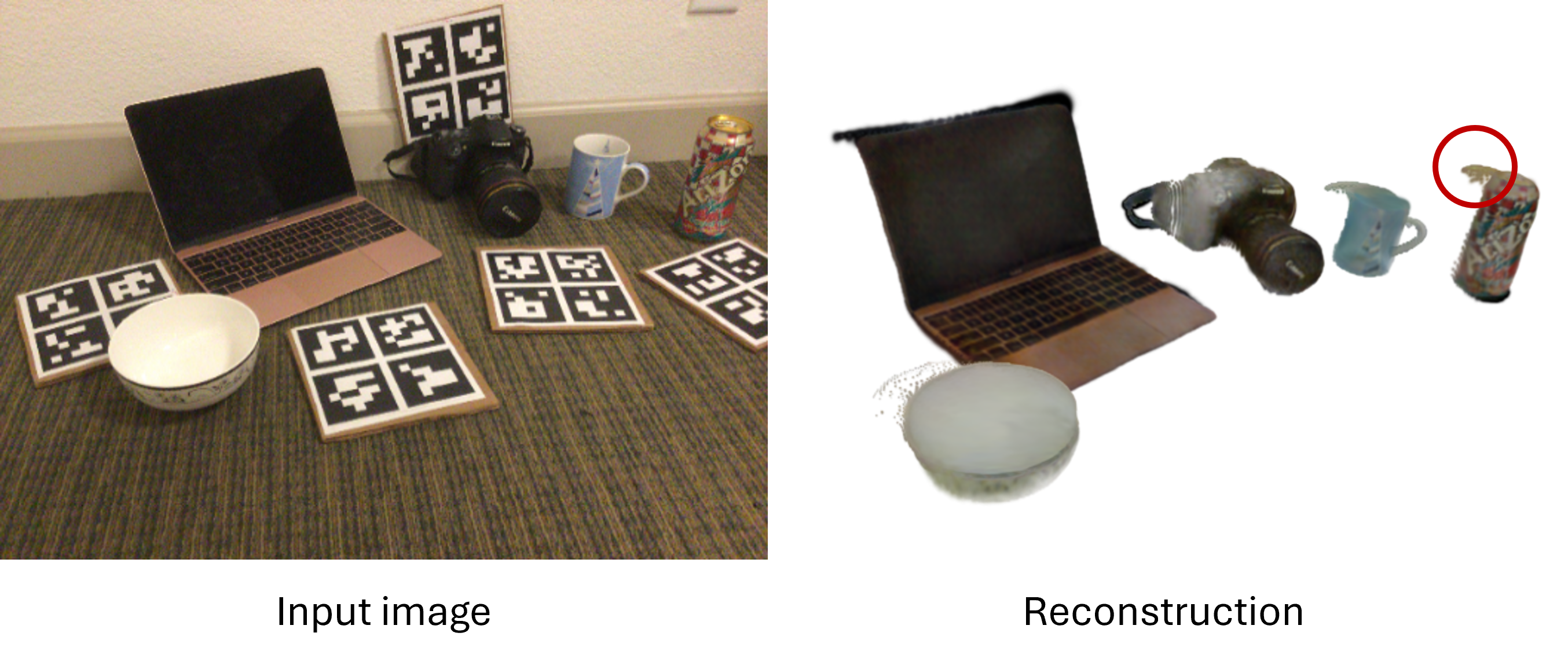}
    \caption{\textbf{Depth--Gaussian coupling.} When the predicted depth is biased, the corresponding Gaussians (circled) are consistently displaced from the true object surface, illustrating the coupling between depth accuracy and 3D reconstruction quality.}
    \label{fig:failure}
\end{figure}

Despite its strong performance, \method has several limitations.
\begin{itemize}
    \item \textbf{Mirror and glass}: Highly reflective or transparent surfaces remain challenging for our depth and NOCS estimators, often producing unstable geometry and inaccurate poses because appearance cues are inconsistent with the underlying 3D structure.
    
    \item \textbf{Depth--Gaussian coupling}: As illustrated in \cref{fig:failure}, Gaussian means inherit their initial placement from the predicted depth before applying the learned off-ray offset. Small, structured depth inaccuracies (e.g., at thin object boundaries) can therefore propagate into mild Gaussian misalignment. This coupling arises only when depth deviates systematically and does not impact most objects.
    
    \item \textbf{Back-side blur}: Because \method relies on a single feed-forward pass without iterative probabilistic refinement (e.g., flow matching or diffusion), rarely observed back surfaces may appear slightly over-smoothed. These cases occur primarily for objects with unusual topology or very uncommon categories where canonical priors are weak (e.g., highly concave shapes, irregular sculptures).
\end{itemize}

\begin{table*}[ht!]
\centering
\small
\setlength{\tabcolsep}{2pt}
\begin{tabular}{lllll}
\toprule
adhesive\_tape\_roll & backpack & bag & ball & bamboo\_mat \\
bamboo\_shoots & banana & bannister & barrel & basket\_container \\
bathtub & battery & bed & bell & belt \\
blinds & board\_game & book & bookshelf & bottle \\
bowl & box & bread & brush & building\_blocks \\
bumbag & burger & cabinet & cable & cake \\
calculator & camera & can & candle & candy \\
carrot & carton\_beverage\_container & ceiling & chair & chess \\
chicken\_leg & chili & chinese\_chess & chip\_can & clip \\
clock & clothes & coconut & column & conch \\
corn & counter & cradle & cucumber & cup \\
curtain & cutter & desk & desk\_caddy & dice \\
dinosaur & dishwasher & doll & donut & door \\
dresser & dumbbell & dumpling & dustpan & egg \\
egg\_tart & electronic\_device & eraser & fan & figure \\
fire\_extinguisher & fireplace & flash\_light & floor & floormat \\
flower\_pot & flute & fork & frisbee & garlic \\
glasses & glasses\_case & gloves & guitar & hair\_dryer \\
hair\_styling\_tool & hairpin & hammer & handbag & hat \\
headset & helmet & hook & hot\_dog & insole \\
instant\_noodle\_cup & kettle & keyboard & kitchen\_utensil & knife \\
lamp & laptop & laundry\_detergent & lemon & lens \\
light & lipstick & magnet & magnifying\_glass & mango \\
mangosteen & marker & medicine\_bottle & melon & mirror \\
monitor & mooncake & mosaic\_tiles & mouse & mug \\
nightstand & notebook & onion & orange & other \\
otherfurniture & otherprop & otherstructure & oven & pad \\
painting & pan & paper & peach & pear \\
pen & pencil & pencil\_case & person & pillow \\
pineapple & pitaya & pitcher & pizza & plate \\
plug & plush\_toy & pomegranate & pot & power\_strip \\
projector & pumpkin & radiator & razor & refrigerator \\
remote\_control & ricecooker & rubik\_cube & sausage & scissor \\
sculpture & shelves & shoe & shower & showercurtain \\
shrimp & sink & slipper & snack & soap \\
soap\_dish & sofa & spanner & speaker & sponge \\
spoon & squeegee & stairs & starfish & stool \\
stove & suitcase & sweet\_potato & sword\_bean & table \\
table\_tennis\_bat & tableware\_set & tape\_measure & teapot & television \\
timer & tissue & toaster & toilet & toiletries \\
tomato & tool & toolkit & toothbrush & towel \\
toy & toy\_animals & toy\_blocks & toy\_boat & toy\_bus \\
toy\_car & toy\_food & toy\_furniture\_set & toy\_motorcycle & toy\_plane \\
toy\_plant & toy\_stacking\_rings & toy\_train & toy\_train\_track & toy\_truck \\
umbrella & usb\_drive & vase & vegetable & waffle \\
wall & wallet & watch & watermelon & whistle \\
whiteboard & window & wine\_glass & wristband &  \\
\bottomrule
\end{tabular}
\caption{List of all 234 semantic categories used in our method.}
\label{tab:categories}
\end{table*}

\FloatBarrier
{
    \small
    \bibliographystyle{ieeenat_fullname}
    \bibliography{main}
}